\documentclass{article}

\usepackage[preprint]{neurips_2026}
\setcitestyle{numbers,square,comma,compress}
\usepackage[utf8]{inputenc} 
\usepackage[T1]{fontenc}    
\usepackage{hyperref}       
\usepackage{url}            
\usepackage{booktabs}       
\usepackage{amsfonts}       
\usepackage{amsmath}
\usepackage{nicefrac}       
\usepackage{cleveref}
\usepackage{microtype}      
\usepackage{xcolor}         
\usepackage{pifont}
\usepackage{graphicx} 
\usepackage{longtable}
\usepackage{tcolorbox}
\usepackage{ragged2e}
\usepackage{multirow} 
\usepackage{adjustbox}
\usepackage{array}    
\usepackage{lscape}   

\usepackage{subcaption}
\usepackage{makecell}
\usepackage{diagbox}
\usepackage{xcolor, colortbl}
\usepackage{caption}
\usepackage{titlesec}
\titlespacing*{\section}{0pt}{*1.0}{*0.1}
\titlespacing*{\subsection}{0pt}{*0.5}{*0.1}
\titlespacing*{\subsubsection}{0pt}{*0.5}{*0.1}
\newcommand{\pp}{\,\mathrm{pp}}
\usepackage{enumitem}

\DeclareRobustCommand{\eg}{\textit{e.g.}}
\DeclareRobustCommand{\ie}{\textit{i.e.}}
\DeclareRobustCommand{\bench}{HumanMoveVQA}

\definecolor{lightgreen}{RGB}{204, 255, 204} 
\definecolor{lightorange}{RGB}{255, 229, 204} 
\definecolor{lightblue}{RGB}{204, 229, 255} 
\newcommand{\best}[1]{\cellcolor{lightgreen} #1} 
\newcommand{\secondbest}[1]{\cellcolor{lightorange} #1} %
\newcommand{\cmark}{\textcolor{green!70!black}{\ding{51}}}
\newcommand{\xmark}{\textcolor{red!70!black}{\ding{55}}}

\newcommand{\thirdbest}[1]{\cellcolor{lightblue} #1} %

\title{HumanMoveVQA: Can Video MLLMs reason about human movement in videos?}

\author{Pulkit Gera$^{1}$ \quad Faegheh Sardari$^{1}$ \quad Asmar Nadeem$^{1}$ \quad Valentina Bono$^{2}$ \And Padraig Boulton$^{2}$ \quad Adrian Hilton$^{1}$ \quad Armin Mustafa$^{1}$
}

\begin{document}

\maketitle

\begin{center}
\vspace{-2em}
\small
$^{1}$ CVSSP, University of Surrey, Guildford, UK\\
$^{2}$ Tesco, UK
\end{center}

\begin{center}
\includegraphics[width=1\textwidth]{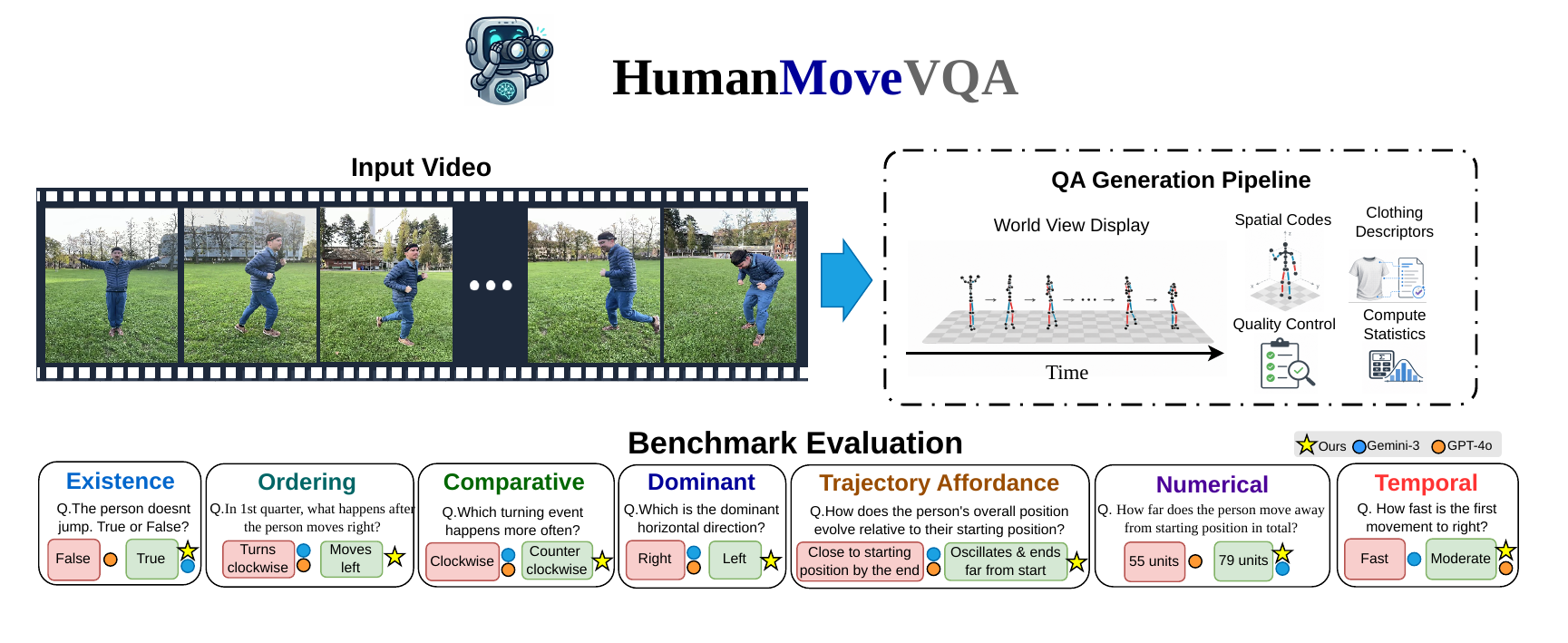}

\captionof{figure}{
Overview of \bench. We evaluate the ability of VideoMLLMs to reason about human movement in videos. Starting from input videos, we extract 3D human pose tracks in a world coordinate system. These motion tracks serve as structured annotations to generate multiple-choice questions across seven categories: Existence, Ordering, Comparative, Dominant, Trajectory Affordance, Numerical, and Temporal. The qualitative comparisons of our model with GPT4o, Gemini-3 shows that our model outperforms in reasoning against closed-source models. Green denotes correct answers and red denotes incorrect ones.}
\label{fig:teaser}
\end{center}
\vspace{-2mm}
\begin{abstract}
\vspace{-3mm}
Despite the rapid advance of Multimodal Large Language Models (MLLMs) in high-level video understanding, a fundamental bottleneck remains: these models collapse complex human motion into coarse semantic labels. Existing benchmarks mostly focus on scene-centric events or local joint articulations, failing to probe global human motion in space over time (trajectory and orientation changes).  
We introduce {\bench}, the first comprehensive benchmark designed to evaluate global trajectory and orientation reasoning from an exocentric perspective. Our benchmark utilizes a first-frame anchored world coordinate system, preserving translation and rotation relative to a fixed starting point. We propose a scalable, multi-stage pipeline that lifts 2D video observations into world-consistent 3D motion tracks to generate over 10K structured question-answer pairs across seven reasoning categories, including motion aggregation, sequential ordering, and trajectory-level inference.  
Our extensive evaluation reveals a critical capability gap in state-of-the-art proprietary models on deep human motion understanding. However, we demonstrate that this is a learnable problem; by fine-tuning an open-source baseline with our targeted, world-consistent supervision, we achieve a significant improvement. {\bench} establishes a rigorous geometric foundation for developing next-generation, movement-aware video understanding models.

\end{abstract}

\section{Introduction}

A simple video caption like \textit{“a person playing tennis”} collapses complex physical sequences into coarse semantic labels, obscuring the underlying global motion dynamics. A rapid lateral sprint to recover a ball is fundamentally different from a slow approach to the net, yet both are unified under the same high-level tag. Crucially, the foundational components of physical action, \textbf{where a person moves, how their trajectory evolves, and how their orientation changes over time} are largely absent from current video-language datasets. This creates a significant information bottleneck for Multimodal Large Language Models (MLLMs), hindering their application in domains like sports analytics, autonomous navigation, and industrial safety.

While recent MLLMs~\citep{gpt4o, Qwen3VL, internvl35} excel at high-level video tasks, such as question answering~\citep{ActivityNetQA,videomme}, causal reasoning \cite{NextQA,star}, video description \citep{MSRVTT}, long-form understanding \cite{longvideobench,lvbench}, and 3D scene understanding \citep{sqa3d}; they are predominantly trained on video-text pairs~\citep{vidal, webvid} with captions that provide only high-level event descriptions. Consequently, current models lack the capacity to reason about a person’s trajectory or orientation changes throughout a sequence. Recent efforts to bridge this gap have focused on fine-grained joint movements~\citep{ActionArt, MMHU, motionllm, motionbench}; however, these address local articulations rather than global motion through space. 
Furthermore, existing benchmarks \cite{ActivityNetQA,MMHU, motionllm, motionbench} are often limited to short-duration clips, precluding the evaluation of long-horizon trajectory and orientation evolution. Parallel research in spatial reasoning~\citep{SpatialReasoner, SpatialVLM, SpatialThinker} focuses on object-level relationships in static images but fails to account for the dynamic, temporal evolution of human movement. This leaves a fundamental question unanswered: Can Video MLLMs reason about human trajectory and orientation in space over time?

In this paper, we introduce \textbf{\bench}, the first comprehensive benchmark for evaluating global trajectory and orientation reasoning in human motion in videos. Unlike previous works that focus on isolated poses, {\bench} requires models to understand movement within a first-frame anchored coordinate system, preserving both translation (displacement) and rotation (orientation) relative to a fixed starting point. We define seven question categories designed to test complementary global movement capabilities, including motion aggregation ({\eg}, counting directional changes), sequence reasoning ({\eg}, temporal ordering of spatial events), and trajectory-level inference ({\eg}, reasoning about displacement or orientation).

To construct {\bench} at scale, we propose a novel pipeline that leverages human mesh reconstruction to recover 3D motion tracks from video, which are then used to generate structured, world-consistent question-answer pairs. This approach enables us to capture continuous human movement in a consistent 3D space over extended sequences, a capability missing from existing benchmarks. We utilise diverse source data from EMDB~\citep{emdb}, RICH~\citep{rich}, and EgoBody~\citep{egobody} to ensure a wide range of movements and environments.

Our evaluation of state-of-the-art Video MLLMs reveals that even capable closed-source models, such as Gemini-3-Flash, achieves an average chance-normalised score of 14.3 across three datasets on trajectory tasks in a zero shot setting.
However, we demonstrate that this is not an architectural ceiling; by fine-tuning an open-source QwenVL3-8B on our generated data, we improve chance-normalised scores three-fold to 43.0. These results suggest that models can learn trajectory- and orientation-level reasoning when provided with targeted, world-consistent supervision.
Our contributions are: 
\begin{itemize}[topsep=0pt,partopsep=0pt,itemsep=0pt,parsep=0pt]
    \item A novel benchmark, {\bench}, evaluating global human trajectory and orientation reasoning using a first-frame anchored coordinate system.
    \item A new pipeline for synthesizing human movement question-answer pairs from world-consistent motion representations for long sequences.
    \item An extensive evaluation demonstrating that state-of-the-art Video MLLMs can significantly improve trajectory and orientation reasoning when fine-tuned on our benchmark.
\end{itemize}

\section{Related Work}
\subsection{Multimodal Large Language Models (MLLMs)}

\begin{table*}[t]
\centering
\caption{Comparison of {\bench} with state-of-the-art benchmarks across key capability axes: \textbf{Video} input, \textbf{Human}-centric, uses an \textbf{Exo}centric viewpoint, evaluates \textbf{Spatial} reasoning grounded in the scene, \textbf{Numerical} reasoning (counting or measuring), \textbf{Trajectory}/path affordance, temporal \textbf{Event Ordering}, and \textbf{Directional} reasoning about movement. {\bench} is the first benchmark to jointly evaluate all axes for human spatial movement from exocentric video.}
\label{tab:benchmark_comparison}
\vspace{2mm}
\resizebox{\textwidth}{!}{
\begin{tabular}{l c c c c c c c c c c}
\toprule
\textbf{Benchmark} & \textbf{Video} & \textbf{Human} & \textbf{Exo} & \textbf{Spatial} & \textbf{Numerical} & \textbf{Trajectory} & \textbf{Event} & \textbf{Directional} & \textbf{\#QA} & \textbf{\#Axes} \\
 & & \textbf{-Centric} & \textbf{-centric} & \textbf{Reasoning} & & \textbf{/Path} & \textbf{Ordering} & \textbf{Reasoning} & \textbf{Pairs} & \\
\specialrule{.2em}{.1em}{.1em}
\multicolumn{11}{c}{\textit{General Video Understanding}} \\
\specialrule{.2em}{.1em}{.1em}
STAR \citep{star} & \cmark & \cmark & \cmark & \xmark & \xmark & \xmark & \xmark & \xmark & 60K & 4 \\
MVBench \citep{mvbench} & \cmark & \xmark & \cmark & \xmark & \xmark & \xmark & \xmark & \xmark & 4K & 20 \\
Video-MME \citep{videomme} & \cmark & \xmark & \cmark & \xmark & \xmark & \xmark & \xmark & \xmark & 2.7K & 12 \\
\specialrule{.2em}{.1em}{.1em}
\multicolumn{11}{c}{\textit{Spatial Reasoning}} \\
\specialrule{.2em}{.1em}{.1em}
SpatialRGPT \citep{SpatialRGPT} & \xmark & \xmark & \cmark & \cmark & \xmark & \xmark & \xmark & \xmark & 1.4K & 12 \\
VSI-Bench \citep{vsibench} & \cmark & \xmark & \xmark & \cmark & \xmark & \cmark & \xmark & \cmark & 5K+ & 8 \\
4D-RGPT \citep{4drgpt} & \cmark & \xmark & \cmark & \cmark & \xmark & \xmark & \xmark & \xmark & 1.5K & 9 \\
SAW-Bench \citep{sawbench} & \cmark & \cmark & \xmark & \cmark & \xmark & \cmark & \xmark & \cmark & 2K+ & 6 \\
\specialrule{.2em}{.1em}{.1em}
\multicolumn{11}{c}{\textit{Human Motion Understanding}} \\
\specialrule{.2em}{.1em}{.1em}
ActionArt \citep{ActionArt} & \cmark & \cmark & \cmark & \xmark & \cmark & \xmark & \xmark & \xmark & 2.7K & 7 \\
MotionBench \citep{motionbench} & \cmark & \xmark & \cmark & \xmark & \cmark & \xmark & \cmark & \xmark & 8K & 6 \\
MotionLLM \citep{motionllm} & \cmark & \cmark & \cmark & \xmark & \xmark & \xmark & \xmark & \xmark & 1.4K & 5 \\
MMHU \citep{MMHU} & \cmark & \cmark & \cmark & \xmark & \xmark & \cmark & \xmark & \xmark & 840 & 13 \\
\specialrule{.2em}{.1em}{.1em}
\textbf{{\bench} (Ours)} & \cmark & \cmark & \cmark & \cmark & \cmark & \cmark & \cmark & \cmark & \textbf{10K} & \textbf{7} \\
\specialrule{.2em}{.1em}{.1em}
\end{tabular}
}
\end{table*}

Recent MLLMs, including VideoLLaVA~\citep{VideoLLava}, VideoGPT+~\citep{VideoGPTplus},InternVL3.5 ~\citep{internvl35} and LLaVA-NeXT-Video~\citep{llava-video}, have achieved significant results in general video understanding. While these models leverage large-scale image-text and video-text datasets~\citep{llava-data, webvid}, such data sources are often insufficient for complex human motion reasoning. Image-text pairs lack temporal information, while video-text captions focus on high-level semantics ({\eg} "a person walks across the room"), 
lacking details on distance, trajectory, or orientation. As a result, these models tend to associate visual patterns with action labels rather than learn trajectory-level spatial reasoning.

To bridge this gap, MotionLLM~\citep{motionllm} and subsequent works~\citep{motionbench, MotionBank, MMHU} have introduced motion encoders and fine-grained motion descriptions to enhance joint-level human reasoning. Similarly approaches like PoseScript ~\citep{PoseScript} and ChatPose~\citep{Chatpose} align SMPL poses and text.However, these methods operate within a canonical coordinate frame that normalises global translation and rotation. While effective for body articulation, they fail to capture a subject’s absolute position and orientation changes over time. Our work addresses this specific gap by focusing on the trajectory and orientation components of human motion that canonical-frame approaches inherently ignore.

\subsection{Video MLLM Benchmarks}

Existing Video MLLM benchmarks predominantly emphasize high-level semantic understanding, spanning tasks such as video question answering \citep{ActivityNetQA}, text-to-video retrieval \citep{howto100m}, and causal or temporal reasoning \citep{NextQA, star, mvbench}. While recent efforts like Video-MME \citep{videomme} and LongVideoBench \citep{longvideobench} extend these evaluations to multimodal and long-form content, they largely treat videos as sequences of discrete semantic events. Consequently, they fail to probe fine-grained human motion or the continuous evolution of spatial trajectories.

Spatial reasoning research has matured in the context of static images \citep{SpatialVLM, SpatialRGPT, SpatialReasoner}, focusing on object-level relative positions. While recent works have transitioned toward spatio-temporal (4D) reasoning \citep{mllm4d, vsibench, 4drgpt}, these remain largely scene-centric. For instance, VSI-Bench \citep{vsibench} and SAW-Bench \citep{sawbench} evaluate route planning from egocentric perspectives. While critical for navigation, these benchmarks do not address the challenge of reasoning about a subject's trajectory and orientation from an exocentric (third-person) viewpoint.

In the human motion domain, benchmarks such as MotionBench \citep{motionbench}, MotionLLM \citep{motionllm}, and MMHU \citep{MMHU} primarily evaluate joint-level articulation. Although MMHU \citep{MMHU} adopts an exocentric perspective and provides pedestrian trajectory annotations, it primarily focuses on behavior classification (e.g., crossing, waiting) rather than structured reasoning about displacement and orientation. The most closely related work, ActionArt \citep{ActionArt}, provides trajectory-aware annotations but prioritizes joint-level understanding over structured reasoning about displacement and orientation. Furthermore, existing motion datasets~\citep{motion-x, motionbench,motion-x} are often limited to short clips ({\ie}, <10 secs), whereas {\bench} captures how movement unfolds over extended temporal periods ({\ie}, 20--60 secs).

As summarized in Table \ref{tab:benchmark_comparison}, {\bench} fills this critical gap. To the best of our knowledge, it is the first benchmark to evaluate human trajectory and orientation from an exocentric perspective over long sequences. By utilising a first-frame anchored coordinate system and long sequences, {\bench} challenges models to move beyond simple high-level semantic action labels toward a rigorous understanding of continuous physical movement.

\section{{\bench}}
\bench\ is designed to evaluate and enhance the capability of MLLMs to reason over human trajectories and orientations within a first-frame anchored world coordinate system. 
Unlike standard Video Question Answering (VideoQA) tasks that often rely on coarse semantic labels like "running," our benchmark requires models to process precise trajectory and orientation transformations relative to the video’s initial frame.
By prioritizing global displacement over local body articulation, we isolate the specific challenge of how a subject traverses and orients themselves within 3D space over time.
To achieve this, we construct multiple-choice questions and answers from deterministically derived world-space motion tracks, enabling scalable generation with verifiable ground-truth evidence. Models must track how a person moves and turns over time, aggregate these changes, and distinguish them from geometrically inconsistent alternatives or global camera movement.  

The benchmark comprises multiple-choice questions across seven distinct categories, targeting three core cognitive axes: (1) Motion Aggregation, (2) Sequential Ordering, and (3) Trajectory-level inference. To ensure that performance reflects true visual reasoning rather than linguistic bias, we propose a "distractor" design. Incorrect options are crafted to be semantically plausible but geometrically inconsistent with the video. This prevents models from succeeding through language priors, speech patterns, or artifacts resulting from the automated template-based generation process. 

The following sections detail our methodology: Section~\ref{sec:data_source} describes the diverse datasets used; Section~\ref{sec:pipeline} outlines our scalable pipeline for extracting world-consistent motion tracks; and Section~\ref{sec:qa_gen} details the deterministic generation of Question-Answer (QA) pairs from these representations.

\subsection{Data Source}
\label{sec:data_source}

We construct \bench\ by leveraging three high-quality human motion datasets: EMDB~\citep{emdb}, EgoBody~\citep{egobody}, and RICH~\citep{rich}. These datasets provide raw video paired with diverse human motion data ({\eg}, SMPL parameters) across a wide range of environments and camera viewpoints. Because these sources utilize disparate coordinate conventions, we process all sequences through a unified multi-stage pipeline (see Section~\ref{sec:pipeline}) to derive consistent, 3D world-space motion tracks.

\textbf{EMDB} consists of monocular videos of a single person, captured with a dynamic handheld camera in indoor and outdoor scenes. Videos are 20–60 secs with substantial variation in motion and camera trajectories, making them suitable for evaluating long-horizon reasoning under complex ego-motion.

\textbf{RICH} contains multi-view videos of a single person performing actions in indoor and outdoor environments. The majority of clips are short (<20 secs) and consist of fewer motion events, providing controlled settings for short-range motion reasoning.

\textbf{EgoBody} contains multi-view (three exocentric viewpoints) two people indoor recordings (15 seconds - several minutes). We segment long videos in 30–60 secs clips and focus on single-person by selecting one individual per sample and include appearance-based descriptors to disambiguate the target person.

For multi-view datasets, RICH and EgoBody, we treat each camera view as an independent sample, increasing data diversity while preserving the underlying motion.


\begin{figure}[t]
    \centering
    \includegraphics[width=\linewidth]{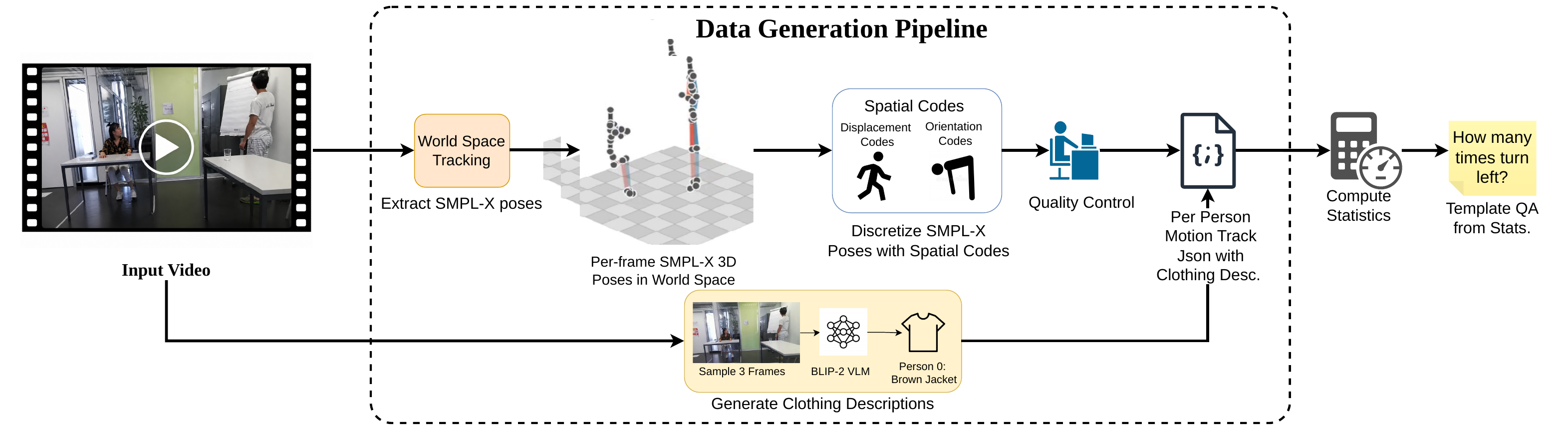}
    \caption{ Overview of pipeline generating \bench .
    Given an input video, we use PromptHMR~\cite{PromptHMR} to recover 3D SMPL-X human poses. We convert these poses into spatial codes capturing root translation and body orientation using MotionScript~\cite{MotionScript}. In parallel, we generate clothing-based descriptors for each person using BLIP-2~\cite{blip2} to enable appearance grounding. We combine spatial codes and descriptors to construct structured motion tracks, which are filtered to remove noisy or unreliable segments. Finally, the verified trajectories are processed to compute motion statistics (e.g., displacement, directional shifts), which are fed into logic-driven linguistic templates to deterministically generate the final question-answer pairs across seven reasoning categories. 
    }
    \vspace{2mm}
    \label{fig:dataset}
\end{figure}
\subsection{Data Generation Pipeline}
\label{sec:pipeline}

To bridge the gap between raw video pixels and symbolic global motion reasoning, we propose a multi-stage pipeline that transforms unstructured video into world-consistent, discretized motion tracks. As illustrated in Figure~\ref{fig:dataset}, the pipeline consists of four key stages: (1) World-Space Lifting, to decouple subject movement from camera motion; (2) Spatial Discretization, to convert continuous signals into reasonable symbolic units; (3) Identity Grounding, to ensure unambiguous reference in multi-person scenes; and (4) Quality Control, to filter noise. This design ensures that resulting QA pairs are grounded in verifiable physical evidence rather than high-level semantic proxies.  

\noindent \textbf{World-Space Reconstruction.}
The primary challenge in exocentric trajectory reasoning is distinguishing between subject movement and ego-motion (camera movement). We use \texttt{PromptHMR}~\cite{PromptHMR} to lift 2D video into 3D SMPL-X~\cite{smplx} human poses within a global coordinate frame. 
By recovering the root translation and orientation in world space, we establish a fixed reference system where all movement is relative to the scene's first frame ($t_0$), neutralizing camera motion or viewpoint. We use a person-centric coordinate system with Y pointing upward, X pointing left-right relative to the subject, and Z pointing forward-backward relative to the subject. 

\noindent \textbf{Discretised Spatial Codes.}
Raw 3D coordinates are often too high-dimensional and noisy for direct language mapping. Inspired by {MotionScript}~\cite{MotionScript}, which represents motion through structured 'MotionCodes' capturing both displacement and rotation, we transform continuous SMPL-X tracks into a discrete set of \textit{Spatial Codes}. We focus on root-level dynamics and define \textit{Spatial Codes} as the subset corresponding to global translation and orientation: \texttt{displacement\_x}, \texttt{displacement\_y}, \texttt{displacement\_z}, \texttt{rotation\_roll}, \texttt{rotation\_pitch}, and \texttt{rotation\_yaw}. 
All values are normalized relative to the first frame to represent change from the initial state. Continuous signals are discretized into categorical bins ({\ie}, displacement: short, moderate, and long; and temporal dynamics: slow and fast). This provides a robust buffer against estimation noise and transforms geometric data into a structured vocabulary suitable for deterministic template-based question generation. Table ~\ref{tab:spatial_codes} and ~\ref{tab:temporal_codes} discuss the spatial and temporal discretization of events in Appendix.

\noindent \textbf{Appearance-Based Identity Grounding.}
In multi-person environments like \textit{EgoBody}, spatial queries must be anchored to a specific person. We use BLIP-2~\cite{blip2} to generate descriptive appearance-based tags (e.g., "blue jacket") by sampling three random frames.
These descriptors are injected into the question prompt to resolve identity ambiguity. This ensures the MLLM's performance measures its spatial tracking ability rather than its ability to guess the target subject.  Figure \ref{fig:cloth_prompt} in Appendix shows the prompt used for generating descriptions.
 
\noindent \textbf{Quality Control.}
To maintain the integrity of the benchmark, we apply a two-tier filtering process. At the video level, we manually remove samples where 3D reconstruction visibly diverges from the 2D video, such as ``drifting'' floor planes or inaccurate tracking. At the event level, we discard motion segments shorter than 5 frames or those categorized as \texttt{very\_short}.
By retaining only temporally significant and stable motion signals, we ensure that questions target meaningful human actions rather than sensor noise or reconstruction artifacts.

\vspace{-2mm}
\subsection{Large Scale Question Answer Generation}
\label{sec:qa_gen}
\vspace{-2mm}

To assess the reasoning capabilities of MLLMs over structured representations, we propose a large-scale, deterministic QA generation framework. By directly mapping symbolic \textit{Spatial Codes} and \textit{Clothing Descriptors} to linguistic templates, we bridge the gap between raw motion tracks and natural language reasoning. We define seven question categories, inspired by \citep{ActionArt,mvbench,motionllm,sawbench} as follows:

\noindent \textbf{{1-} Existence:} Evaluates the detection of motion events, such as directional movement or rotation. These binary QA tasks mix positive and negative samples to prevent simple semantic memorization. 

\noindent \textbf{{2-} Comparative:} Probes pairwise reasoning by querying the model to compare opposing directions along the same axis based on frequency, magnitude, or speed. Tests are conducted globally and within temporal quarters to evaluate localized aggregation, comparative judgement and tie handling.

\noindent \textbf{{3-} Dominant:} The video is divided into four temporal quarters. Similar to comparative tasks, this requires threshold-aware aggregation to identify the primary direction or rotation along a specific axis, including handling ties between opposing events and tested both globally and in temporal quarters.

\noindent \textbf{{4-} Numerical:} Tests the MLLM's ability to count discrete motion events or estimate total displacement magnitude over the sequence.

\noindent \textbf{{5-} Ordering:} Tests fine-grained sequential reasoning by requiring the model to identify the event immediately following a specified anchor within a temporal quarter.

\noindent \textbf{{6-} Temporal:} Integrates event detection with temporal localization, asking the model to identify the earliest event in a temporal quarter or categorize the speed of a specific movement.

\noindent \textbf{{7-} Trajectory Affordance:} Challenges the model to reason over long-horizon motion by aggregating displacement across time, including net displacement, total path length, trajectory shape, and quarter-level trajectory understanding.

For each video, we generate up to ten questions per category through a deterministic mapping of motion tracks to linguistic templates, ensuring that every query is grounded in verifiable spatial evidence. To prevent shortcut solutions, we design answer options to be semantically plausible while controlling for structural biases. To ensure \bench\ measures true visual reasoning rather than linguistic heuristics, we employ a rigorous distractor design. For numerical tasks, distractors are sampled within $\pm 3$ of the correct count and $\pm 20\%$ for magnitudes. Comparative and dominant options are restricted to valid opposing directions along the same axis, with explicit tie options, and controlled tie-frequency to avoid bias. Temporal and ordering options contain balanced mixtures of displacement and rotation labels to prevent elimination by motion type. Finally, trajectory affordance options reflect plausible outcomes based on displacement, path, and shape reasoning. 
Across all categories, option positions are shuffled, and answer distributions are constrained to be approximately uniform, ensuring correct answers require reasoning over the underlying spatial structure. Comprehensive examples of these question-answer pairs and their corresponding templates are detailed in Appendix.

\vspace{-3mm}
\subsection{Benchmark Composition and Statistics}
\label{sec:benchmark_design}

We construct the {\bench}  dataset by selecting sequences with high-fidelity tracking and diverse motion profiles. The benchmark is partitioned into training and test sets. The test set is videos with high-quality tracking and diverse motion patterns across displacement and rotation. To ensure rigorous evaluation and prevent data leakage, all camera viewpoints corresponding to a single underlying 3D motion in multi-view datasets (RICH and EgoBody) are assigned to the same split. The resulting composition, summarized in Table~\ref{tab:benchmark_stats}, comprises of 10,203 question-answer pairs. 
The ordering category yields a lower question count relative to other axes across all subsets. This is due to stricter logic-driven constraints: a valid query requires a temporal quarter to contain at least two discrete motion events with spatial category \textit{moderate} or \textit{strong}.

\begin{table}[t]
\centering
\caption{Test set statistics for {\bench}.}
\label{tab:benchmark_stats}
\resizebox{\textwidth}{!}{
\begin{tabular}{lccccccccc}
\specialrule{.2em}{.1em}{.1em}
Dataset & Videos & Total & Existence & Numerical & Comparative & Dominant & Temporal & Ordering & Traj. Afford. \\
\specialrule{.2em}{.1em}{.1em}
EMDB    & 11 & 786   & 143   & 110 & 110 & 110 & 110 & 100 & 93  \\
RICH    & 32 & 2,220 & 416   & 320 & 318 & 320 & 320 & 227 & 299 \\
EgoBody & 54 & 7,197 & 1,378 & 1,060 & 1,044 & 1,046 & 1,053 & 656 & 960 \\
\specialrule{.2em}{.1em}{.1em}
Total   & 97 & 10,203 & 1,937 & 1,490 & 1,472 & 1,476 & 1,483 & 983 & 1,352 \\
\specialrule{.2em}{.1em}{.1em}
\end{tabular}
}
\end{table}

\begin{figure}[t]
\centering
\begin{subfigure}[b]{0.32\textwidth}
    \centering
    \includegraphics[width=\linewidth]{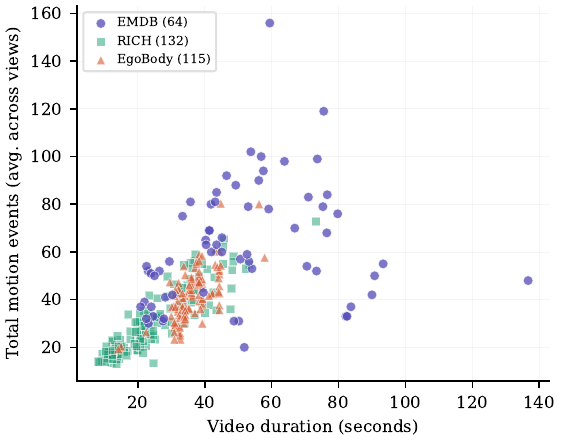}
    \caption{Motion Events vs Video Length}
    \label{fig:vidlength}
\end{subfigure}%
\hfill
\begin{subfigure}[b]{0.32\textwidth}
    \centering
    \includegraphics[width=\linewidth]{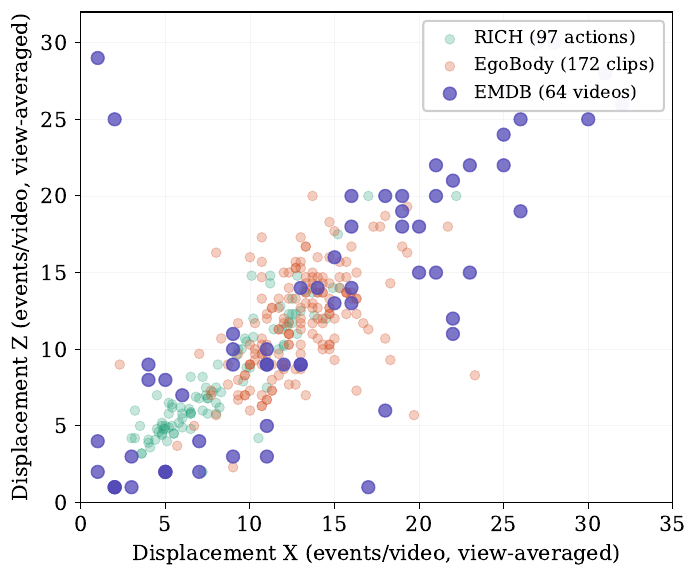}
    \caption{Displacement X vs Z}
    \label{fig:disp_xz}
\end{subfigure}%
\hfill
\begin{subfigure}[b]{0.32\textwidth}
    \centering
    \includegraphics[width=\linewidth]{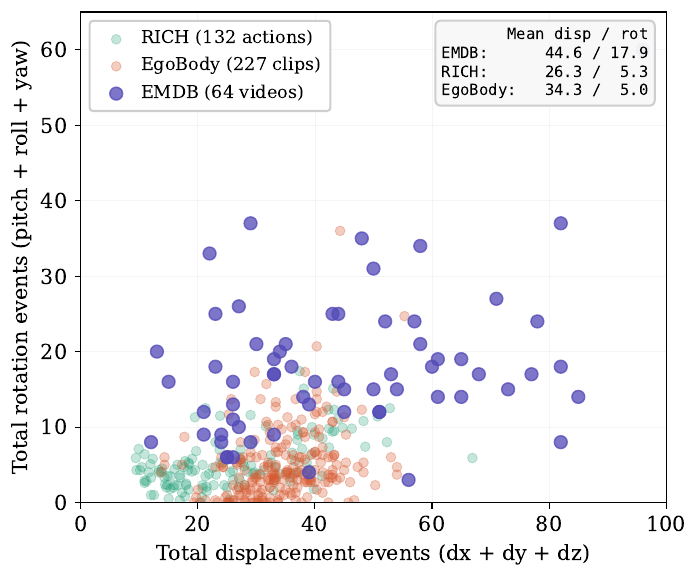}
    \caption{Displacement vs Rotation}
    \label{fig:disp_rot}
\end{subfigure}
\caption{Motion characteristics across the three full datasets before train/test splitting. 
\textbf{Left:} Motion events vs.\ video length, showing that RICH consists of shorter clips with fewer motion events, EgoBody contains longer clips but limited motion activity, and EMDB, despite fewer videos, exhibits substantially higher event counts. 
\textbf{Middle:} Displacement in X vs.\ Z directions per video (view-averaged). 
\textbf{Right:} Total displacement vs.\ total rotation events per video (view-averaged). }
\label{fig:motion_stats}
\end{figure}

Figure~\ref{fig:motion_stats} 
illustrates the overall motion characteristics of {\bench} across its three source datasets. Figure ~\ref{fig:vidlength} visualizes the distribution of motion events and video length. EMDB has the fewest videos with high variance of motion diversity. EgoBody has fewer events due to two people talking and facing each other. RICH has the shorter clips and subsequently fewest motion events. Figure~\ref{fig:disp_xz} shows per-video displacement event counts along the horizontal ($x$) and depth ($z$) axes reveal that EMDB spans the widest range, reflecting its longer temporal horizons and more dynamic motion. Conversely, RICH clusters in the low-event-count region due to its shorter clips and limited movement, while EgoBody exhibits a moderate spread. When comparing total displacement to total rotation (Figure~\ref{fig:disp_rot}), EMDB demonstrates high variance in both metrics. EgoBody shows comparable displacement to EMDB in some instances but consistently lower rotation, indicating movement patterns are mostly translational. RICH remains concentrated in the low-displacement, low-rotation region.
Overall, {\bench} ensures that answering questions requires reasoning over motion trajectories and orientation changes, rather than relying on superficial cues or dataset biases.

\vspace{-2mm}
\section{Experiments}
\subsection{Results}
In this section, we present both quantitative and qualitative results from our proposed benchmark, evaluating various models on the dataset. 
Evaluation setup and implementation details are in Appendix. We provide qualitative video results across all three datasets in the supplementary.


\noindent{\bf{Metrics --}} As the number of multiple-choice options varies by category, random chance thresholds differ: 50\% for Existence (2 options), 33.33\% for Comparative and Dominant (3 options), 25\% for Numerical, Ordering, Temporal and Trajectory Affordance (4 options). For fair comparison across categories, we report the per category accuracy and aggregate \textbf{Score} for all models, defined as the mean chance-normalized accuracy: $\text{Score} = \frac{1}{n} \sum_{i=1}^{n=7}
    \frac{\text{accuracy}_i - \text{chance}_i}{100 - \text{chance}_i}
    \times 100$
%

\begin{figure*}[t]

    \centering

    \includegraphics[width=\textwidth]{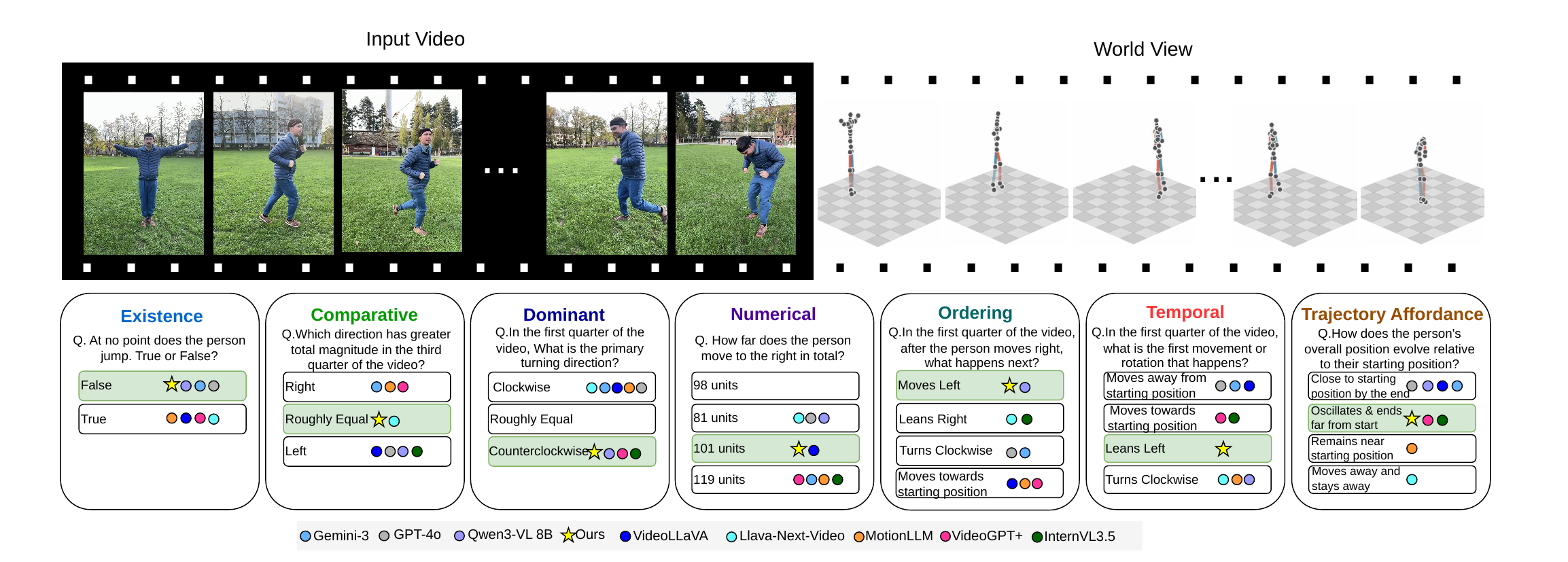}
    \caption{Qualitative results on the \textbf{EMDB} dataset. The world-view depicts extracted 3D SMPL-X poses and is shown for illustration only (not provided to the models). Green denotes the correct option. We compare predictions from multiple MLLMs across seven reasoning categories, where our model demonstrates more accurate and consistent responses. More results in Figure \ref{fig:emdb_2} (appendix)}
    
    \label{fig:emdb}

    \vspace{-2mm}

\end{figure*}

We perform evaluations on all three datasets, EMDB (below) and the RICH and Egobody dataset evaluations are added in the Appendix. 
\begin{table}[b]
\centering
\caption{Evaluation of video MLLM models on the \textbf{EMDB} split. Results are reported across seven reasoning categories and an overall normalized score. \best{Green}, \secondbest{orange}, and \thirdbest{blue} indicate the best, second-best, and third-best results per column, respectively.}
 \label{tab:emdb}
\vspace{2mm}
    \renewcommand{\arraystretch}{1.3}
    \setlength{\tabcolsep}{5pt}
    \resizebox{\textwidth}{!}{
    \begin{tabular}{l|c|c|c|c|c|c|c|c|c}
\specialrule{.2em}{.1em}{.1em}
Model 
& Frames
& Existence 
& Comparative  
& Dominant 
& Numerical 
& Ordering 
& Temporal 
& Traj. Afford.
& Score \\
\specialrule{.2em}{.1em}{.1em}
Random Chance
& --
& 50.00
& 33.33
& 33.33
& 25.00
& 25.00
& 25.00
& 25.00
& 0.00 \\
Human (subset)
& --
& 100.00
& 78.67
& 83.33
& 85.50
& 75.50
& 95.00
& 75.00
& 78.71 \\
\specialrule{.2em}{.1em}{.1em}
\multicolumn{10}{c}{\textit{\bf Proprietary Models}} \\
\specialrule{.2em}{.1em}{.1em}

GPT-4o
& 32
&60.10 &40.00 &40.90 &29.01 &32.00 &29.09 &14.00 &6.84  \\

Gemini-3-flash
& 2 FPS
&\secondbest{}80.42 &30.00 &42.73 &28.18 &29.00 &\secondbest{}43.64 &\secondbest{}32.26 &\secondbest{}16.29  \\

Gemini-3-flash text
& --
&60.14 &32.73 &48.18 & 18.18 &38.00 &29.09 &27.96 & 8.47 \\
\specialrule{.2em}{.1em}{.1em}
\multicolumn{10}{c}{\textit{\bf Open-Source MLLM }} \\
\specialrule{.2em}{.1em}{.1em}

VideoGPT+ 4B
& 16
&53.85 &32.73 &44.04 &25.45 &30.00 &22.94 &25.81 & 4.19 \\

Qwen3-VL 4B
& 32
&\thirdbest{}64.34 &34.55 &\thirdbest46.36 &\secondbest{}31.82 &\secondbest{}38.00 &29.09 &26.96 & 12.26 \\

Video-LLaVA 7B
& 8
&61.54 &\thirdbest{}40.37 &39.09 &27.27 &27.00 &23.64 &17.39 & 5.35 \\

LLaVA-NeXT-Video-7B
& 32
&51.77 &30.00 &43.66 &25.45 &24.00 &28.18 &25.80 & 2.71 \\

InternVL3\_5-8B
& 8
&54.61 &\secondbest{}41.28 &36.36 &27.27 &\thirdbest{}31.00 &\thirdbest{}32.11 &12.09 & 4.00 \\

Qwen3-VL 8B
& 32
&\thirdbest{}64.34 &\best{}43.64 &\secondbest{}49.09 &\thirdbest29.09 &30.00 &29.09 &\thirdbest27.96 & \thirdbest12.83 \\
\specialrule{.2em}{.1em}{.1em}
\multicolumn{10}{c}{\textit{\bf Motion-Specialized Model}} \\

\specialrule{.2em}{.1em}{.1em}
MotionLLM 7B
& 8
&46.15 &28.18 &36.56 &6.36 &17.00 &21.10 &12.90 & -9.53 \\ 

 \midrule \midrule

{\bf Qwen3-VL 8B SFT {(Ours)}}
& 32
& \best{}83.92 & 38.18 & \best{}56.36 & \best{}70.00 & \best{}47.00 & \best{}49.09 & \best{}50.54 & \best{}37.88 \\ 

\specialrule{.2em}{.1em}{.1em}
\end{tabular}
    }
\end{table}
Table \ref{tab:emdb} and Figure \ref{fig:emdb} presents results for EMDB's long-duration ({\ie}, 20–60s), high-diversity monocular videos. Among proprietary models, Gemini-3-Flash leads (Score: 16.3), excelling in Existence (80.4) and Temporal (43.6), but falling below the Random Chance baseline in Comparative (30.0). This indicates a failure to aggregate detected events. GPT-4o struggles with Trajectory Affordance (14.0), failing at long-term tracking. The text-only baseline (Score: 8.5) exceeds GPT-4o, primarily through language biases in the Dominant category.

Among open-source models, Qwen3-VL 8B is the strongest zero-shot baseline (Score: 12.8). Video-LLaVA, VideoGPT+,InternVL 3.5 and LLaVA-Video-NeXT perform marginally above chance. MotionLLM performs worst, confirming that joint-level supervision does not assist in global trajectory reasoning. 
Our Qwen3-VL 8B SFT model nearly triples the base performance (Score: 37.9), with major gains in Numerical ($+40.9\pp$), Trajectory Affordance ($+22.6\pp$), and Ordering ($+17.0\pp$). 



\noindent \textbf{Takeaways:}
The evaluation of state-of-the-art models on \bench\ reveals several critical insights regarding the current state of motion reasoning in Video MLLMs:   

\begin{itemize}[leftmargin=*,topsep=0pt,partopsep=0pt,itemsep=0pt,parsep=0pt]
\item \textbf{Zero-Shot Limitations:} Current zero-shot baselines struggle significantly with motion reasoning in space and time, failing to achieve meaningful performance gains in the \textit{Numerical}, \textit{Ordering}, and \textit{Trajectory Affordance} categories across all evaluated datasets. 
\item \textbf{Leading Baselines:} \textit{Gemini-3-Flash} emerges as the strongest zero-shot performer; however, its success is largely restricted to \textit{Existence} and \textit{Temporal} tasks, as it performs near the random chance baseline in \textit{Comparative} reasoning. 
\item \textbf{Impact of Targeted Supervision:} Our \textit{Qwen3-VL 8B SFT} model significantly advances the state-of-the-art, nearly tripling aggregate performance in some settings and demonstrating that trajectory-level intelligence is a learnable when provided with targeted, world-consistent supervision. 
\item \textbf{Insufficiency of Joint-Level Data:} The results highlight that joint-level motion supervision, as utilized by models like \textit{MotionLLM}, is insufficient for global movement reasoning, as that model consistently performs poorly on trajectory-based tasks. 
\item \textbf{Primary Technical Challenge:} \textit{Ordering} is the most challenging task in the benchmark, requiring the complex detection and precise temporal sequencing of multiple discrete motion events. 
\end{itemize}


\begin{table}[t]
\centering

 \caption{\textbf{Cross-Dataset Evaluation}. In-domain and cross-domain performance of Qwen3-VL 8B when trained on different dataset subsets. Results highlight generalization across datasets; green indicates the best result on the test set, and orange indicates the second best.}
 \label{tab:cross_dataset}
\vspace{2mm}
    
    \renewcommand{\arraystretch}{1.3}
    \setlength{\tabcolsep}{5pt}
    \resizebox{\textwidth}{!}{
    \begin{tabular}{lcccccccc}
\specialrule{.2em}{.1em}{.1em}
Train $\rightarrow$ Test 
& Existence 
& Comparative
& Dominant
& Numerical 
& Ordering  
& Temporal
& Traj. Afford. 
& Score\\
\specialrule{.2em}{.1em}{.1em}
EMDB $\rightarrow$ EMDB

&\best{}88.11 &\secondbest{}38.18 &44.55 &\best{}66.36 &\best{}48.00 &\secondbest{}49.09 &\secondbest{}45.16 &\best{}35.02 \\
RICH $\rightarrow$ EMDB
&\secondbest{}74.13 &\best{}40.00 &\secondbest{}50.91 &63.64 &33.00 &49.09 &39.78 &28.38 \\
EgoBody $\rightarrow$ EMDB
&71.33 &36.36 &\best{}60.91 &\secondbest{}65.45 &\secondbest{}40.00 &\best{}53.64 &\best{}49.46 &\secondbest{}33.33 \\
\specialrule{.2em}{.1em}{.1em}
EMDB $\rightarrow$ RICH
&76.44 &44.97 &49.38 &66.25 &\secondbest{}35.24 &42.50 &\secondbest{}43.81 &30.21 \\
RICH $\rightarrow$ RICH
&\best{}85.34 &\best{}48.74 &\best{}63.75 &\secondbest{}70.94 &\best{}37.44 &\best{}57.81 &\best{}52.17 &\best{}42.46 \\
EgoBody $\rightarrow$ RICH
&\secondbest{}80.53 &\secondbest{}46.54 &\secondbest{}56.56 &\best{}74.69 &32.60 &\secondbest{}52.19 &42.14 &\secondbest{}35.89 \\
\specialrule{.2em}{.1em}{.1em}
EMDB $\rightarrow$ EgoBody
&\secondbest{}75.11 &48.08 &53.92 &\secondbest{}66.98 &\secondbest{}32.47 &41.03 &\secondbest{}43.85 &30.81 \\
RICH $\rightarrow$ EgoBody

&73.59 &\secondbest{}52.07 &\best{}60.76 &66.08 &\best{}33.38 &\secondbest{}52.94 &41.54 &\secondbest{}34.53 \\
EgoBody $\rightarrow$ EgoBody
&\best{}76.92 &\best{}54.79 &\secondbest{}60.04 &\best{}73.58 &32.01 &\best{}55.94 &\best{}49.69 &\best{}39.20 \\
\specialrule{.2em}{.1em}{.1em}
\end{tabular}
    }

\end{table}
\renewcommand{\arraystretch}{1.15}
\setlength{\tabcolsep}{4pt}
\begin{figure}[t]
\centering
\begin{subfigure}[b]{0.40\textwidth}
    \centering
    \includegraphics[width=\linewidth]{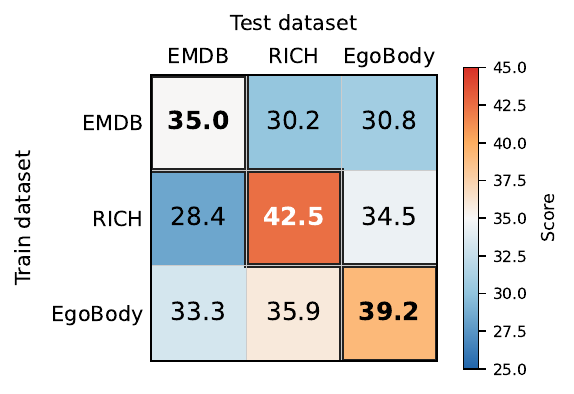}
    \caption{Cross-dataset generalization showing performance when training on one dataset and evaluating on another.}
    \label{fig:crossdomain_heatmap}
\end{subfigure}
\hfill
\begin{subfigure}[b]{0.59\textwidth}
    \centering
    \includegraphics[width=\linewidth]{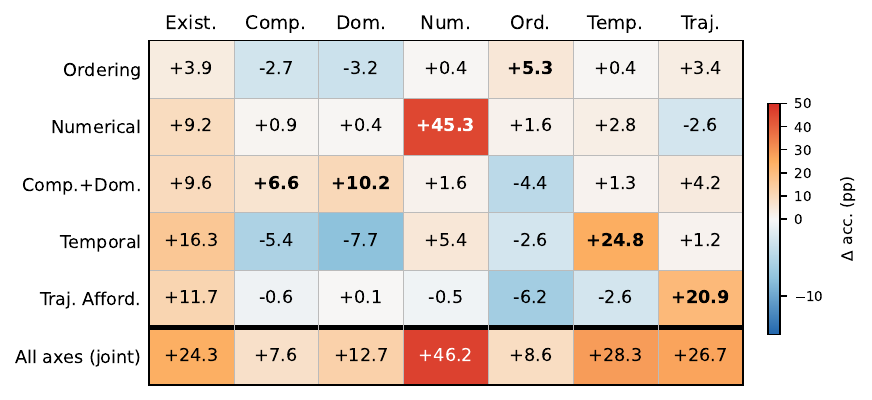}
    \caption{Per-category vs.\ joint SFT training evaluated across task axes, reported as change in accuracy (pp) averaged over datasets.}
    \label{fig:axis_heatmap}
\end{subfigure}

\caption{Heatmap visualizations of training strategies and cross-dataset generalization. 
}
\label{fig:heatmap}
\end{figure}
\vspace{-2mm}
\subsection{Ablations}
We conduct a series of ablations to evaluate the impact of our design choices on supervised fine-tuning (SFT) performance. In this section, we evaluate two questions: How well does training on one dataset transfer to other datasets?; and, Do we need to train on all seven categories to observe gains on all categories? We evaluate four more questions on the effect of resolution, frame count, reasoning supervision and out-of-domain evaluation in the Appendix.

\noindent{\bf{Cross-Dataset Generalization --}} We explore how does training on one dataset generalize to evaluation on other datasets. As shown in Table \ref{tab:cross_dataset}, in-domain training consistently yields the highest Score, reflecting unique motion distributions across datasets. However, the transfer is asymmetric: Figure \ref{fig:crossdomain_heatmap} demonstrates that training on EgoBody generalizes best, matching the in-domain performance of EMDB and RICH. This robustness is likely due to EgoBody’s larger sample size ({\ie}, \textasciitilde10$\times$ EMDB) and complex multi-person scenarios. Conversely, EMDB generalizes the least. Notably, Numerical, Temporal, and Comparative reasoning show stable cross-dataset transfer, while Ordering remains the most challenging category across all domains. These results justify our joint-training strategy, as no single dataset dominates all categories.

\noindent{\bf{Per-Category vs. Joint Training --}} We investigate whether specialized training on a single category can match joint training across all axes. We trained five specialized models (Numerical, Ordering, Temporal, Trajectory Affordance, and a combined Comparative+Dominant model). Table \ref{tab:axis_ablation_all_datasets} in appendix, shows that while category-specific training improves its target axis ({\eg}, Numerical improves by $+45.3pp$), it often degrades others; for instance, Temporal SFT reduces Dominant accuracy by $-7.7 pp$. Existence is the only category that improves universally as a side-effect of any SFT run. 
Figure \ref{fig:axis_heatmap} shows these effects averaged across three datasets. Ordering is the least responsive category, Existence category improves as a side effect in every category and joint training provides the best overall results, demonstrating positive transfer across various categories.

\

\vspace{-7mm}
\section{Conclusion}
We introduced \bench, the first comprehensive benchmark designed to assess the ability of Video MLLMs to reason about global human movement, specifically trajectory and orientation information. By proposing a scalable pipeline that transforms 3D motion tracks into structured, world-consistent question-answer pairs within a first-frame anchored coordinate system, we enable rigorous evaluation across seven reasoning categories: Existence, Comparative, Dominant, Numerical, Ordering, Temporal, and Trajectory Affordance.
Our extensive empirical evaluation reveals a substantial gap in current MLLM capabilities. State-of-the-art models, including GPT-4o and Gemini-3-Flash struggle particularly with counting and long-horizon spatial aggregation. Furthermore, we demonstrate that joint-level motion supervision proposed by MotionLLM is insufficient for global movement reasoning. However, we show that this is a learnable challenge; supervised fine-tuning (SFT) of Qwen3-VL 8B on our data yielded nearly a three-fold improvement in aggregate score.
Ablation studies confirm the effectiveness of our design, showing that joint training across all categories outperforms axis-specific specialists and that learning generalizes effectively across different datasets. 
Finally, our results indicate that motion reasoning is more sensitive to spatial resolution, and that simplified, option-only supervision currently outperforms complex reasoning traces. Overall, {\bench} provides a critical foundation for developing next-generation video models with a robust, geometric understanding of human movement.





\noindent{\bf{Limitations -- }}
The benchmark relies on reliable motion tracking from monocular videos. Noisy or ambiguous annotations, particularly for subtle rotations and short-duration events, may affect both training signal quality
and evaluation reliability. We do not evaluate multi-person movement or interactions between them. Finally our SFT training demonstrated that tasks like Ordering might require some other innovations.

\clearpage
\bibliographystyle{unsrtnat}
\bibliography{references}
\appendix

\section{Appendix}
\subsection{Experiments}
\noindent{\bf{Evaluation Setup --}} We evaluate a diverse suite of models, including proprietary, open-source, and motion-specialized Video MLLMs. For proprietary models, we test GPT-4o~\citep{gpt4o} and Gemini-3-flash~\citep{gemini3flash}, the latter of which is evaluated in both a standard video-text setting and a text-only setting to quantify the reliance on visual information. Open-source baselines include Video-LLaVA~\cite{VideoLLava}, VideoGPT+~\citep{VideoGPTplus}, LLaVA-NeXT-Video~\cite{llava-video},InternVL3.5 ~\cite{internvl35} and Qwen3-VL~\citep{Qwen3VL}. We also evaluate MotionLLM~\citep{motionllm}, which incorporates an explicit motion encoder trained on joint-level captions. All baselines are evaluated zero-shot using recommended configurations. Additionally, we conduct human evaluation on a small subset of the dataset. Participants were given one question of each type per video for evaluation.

\noindent{\bf{Implementation Details --}} We perform supervised fine-tuning (SFT) on Qwen3-VL 8B using our training set of 89,818 QA pairs derived from EMDB (3,958), RICH (45,537), and EgoBody (40,323). Training is conducted via LoRA (rank 16, alpha 32) for 5 epochs using the LlamaFactory ~\cite{llamafactory} framework. We sample 32 uniform frames at 256$\times$256 resolution and utilize early stopping based on a held-out validation set. Training was completed in one day on two NVIDIA H100 GPUs with a batch size of 2. System prompt used for training and evaluation of models is in Figure ~\ref{fig:system_prompt}.

\subsubsection{Results}
\noindent{\textbf{RICH --}} The RICH dataset consists of short clips (10–20s) captured by static cameras, with fewer motion events compared to EMDB. As shown in Table \ref{tab:rich}, Gemini-3-Flash leads zero-shot models with a Score of $16.34$, primarily due to its performance in the Temporal category ($45.31$). GPT-4o lags behind the Qwen3-VL variants and continues to struggle with Trajectory Affordance tasks. The text-only baseline performs near the Random Chance baseline, confirming the absence of exploitable linguistic priors in the benchmark.
\begin{figure*}[b]

    \centering

    \includegraphics[width=\textwidth]{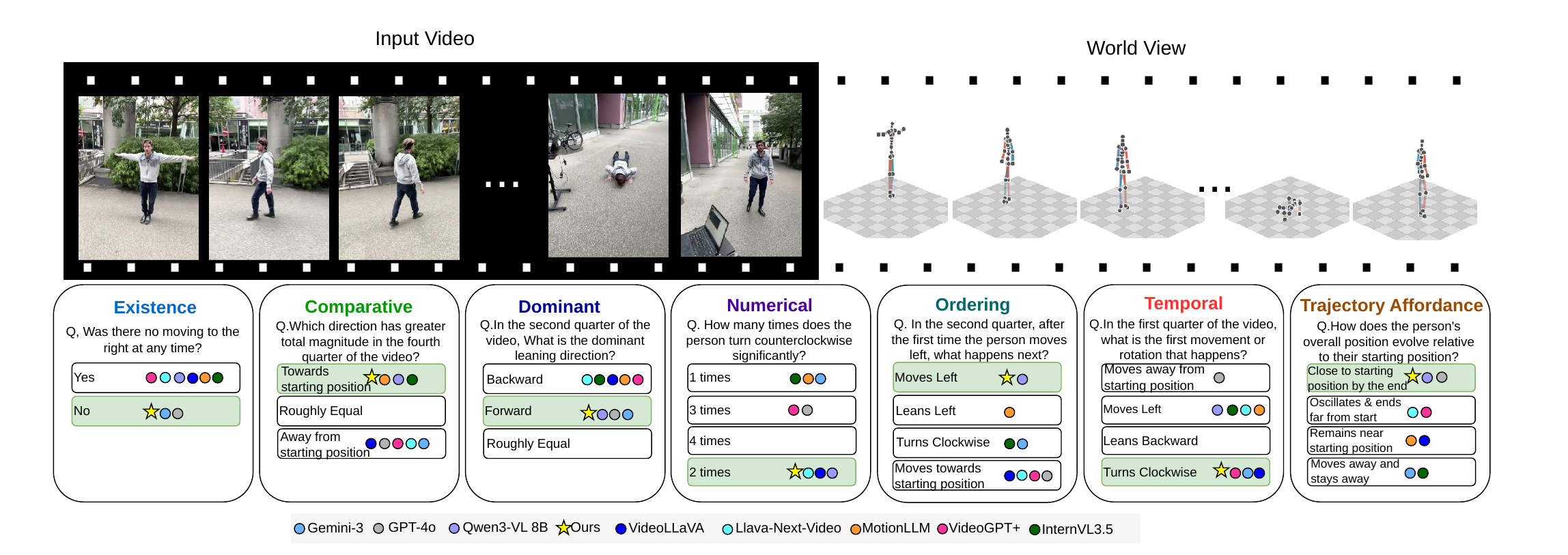}

    \caption{Qualitative results on the \textbf{EMDB} dataset. The world-view visualization depicts extracted 3D SMPL-X poses and is shown for illustration only (not provided to the models). Green denotes the correct option. We compare predictions from multiple MLLMs across seven reasoning categories, where our model demonstrates more accurate and consistent responses.}
    
    \label{fig:emdb_2}


\end{figure*}
\begin{figure*}[t]

    \centering

    \includegraphics[width=\textwidth]{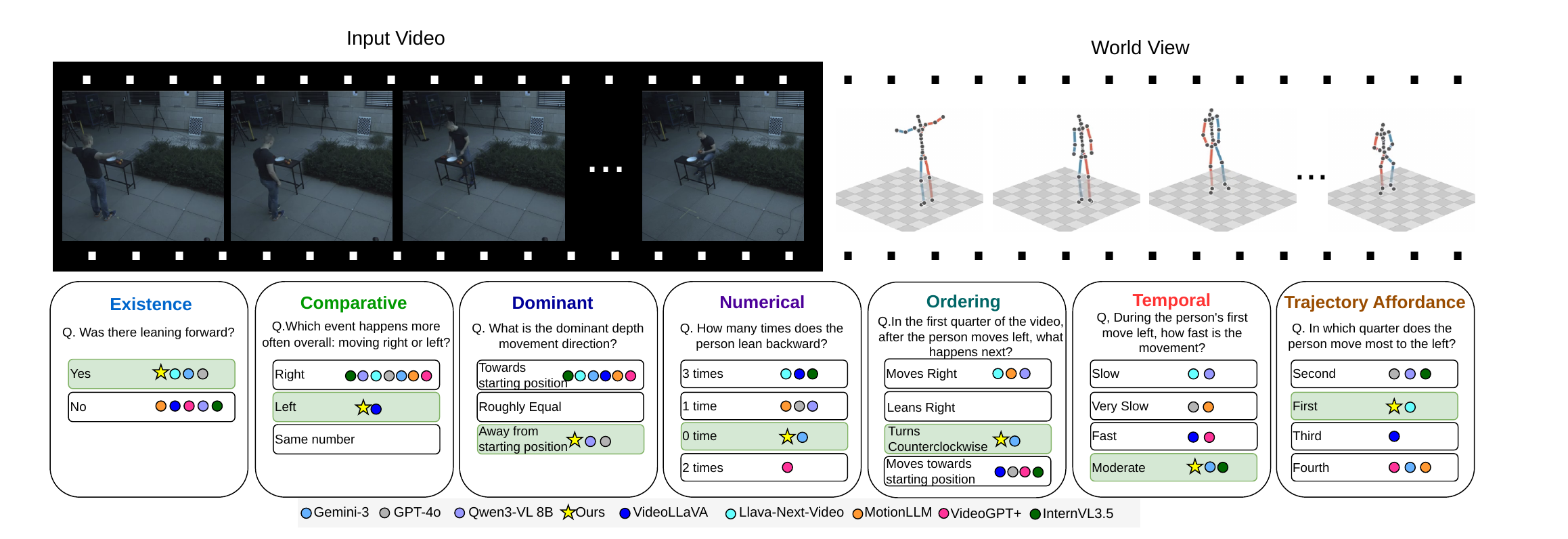}

    \includegraphics[width=\textwidth]{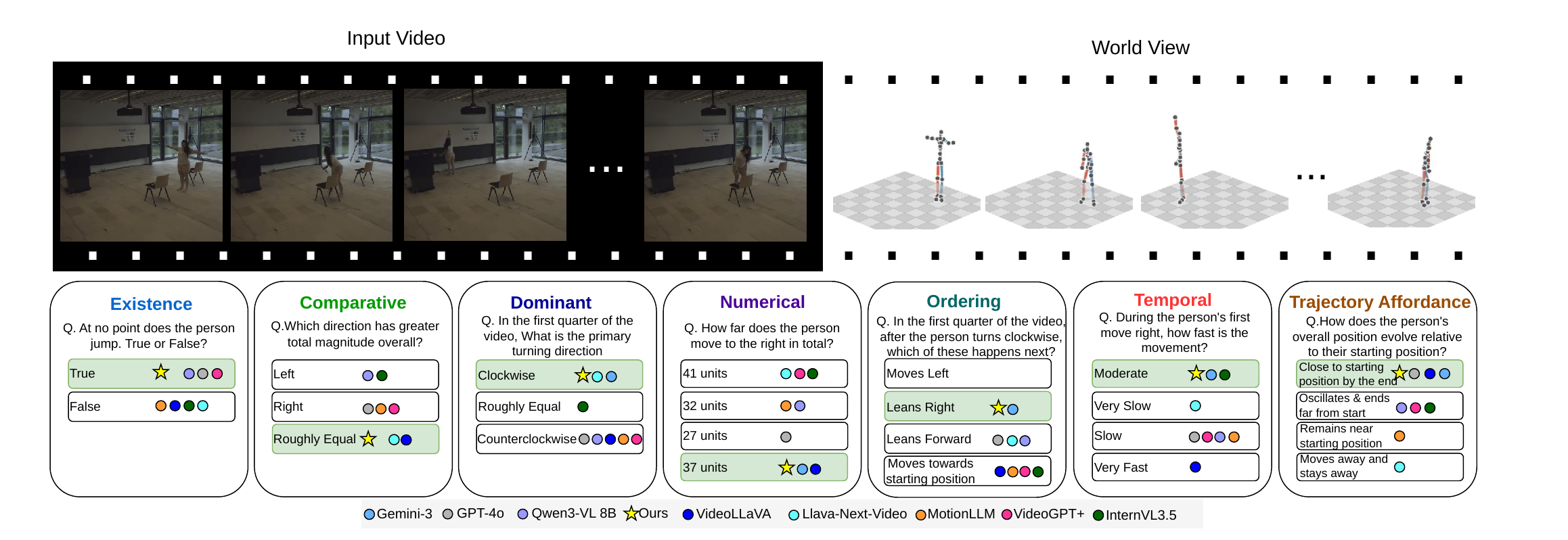}
    
    \caption{Qualitative results on the \textbf{RICH} dataset. The world-view visualization depicts extracted 3D SMPL-X poses and is shown for illustration only (not provided to the models). Green denotes the correct option. We compare predictions from multiple MLLMs across seven reasoning categories, where our model demonstrates more accurate and consistent responses.}
    
    \label{fig:rich}


\end{figure*}
\begin{table}[h]
\centering
\caption{Evaluation of video MLLM models on the \textbf{RICH} split. Results are reported across seven reasoning categories and an overall normalized score. \best{Green}, \secondbest{orange}, and \thirdbest{blue} indicate the best, second-best, and third-best results per column, respectively for MLLMs.}
 \label{tab:rich}
    \renewcommand{\arraystretch}{1.3}
    \setlength{\tabcolsep}{5pt}
    \resizebox{\textwidth}{!}{
    \begin{tabular}{l|c|c|c|c|c|c|c|c|c}
    \specialrule{.2em}{.1em}{.1em}
Model 
& Frames
& Existence 
& Comparative  
& Dominant 
& Numerical 
& Ordering 
& Temporal 
& Traj. Afford.
& Score \\
\specialrule{.2em}{.1em}{.1em}
Random Chance
& --
& 50.00
& 33.33
& 33.33
& 25.00
& 25.00
& 25.00
& 25.00 
& 0.00\\
Human (subset)
& --
& 100.00
& 83.40
& 88.56
& 80.50
& 75.50
& 90.00
& 75.50 
& 79.04\\
\specialrule{.2em}{.1em}{.1em}

\multicolumn{10}{c}{\textit{\bf Proprietary Models}} \\
\specialrule{.2em}{.1em}{.1em}

GPT-4o
& 32
&\thirdbest{}63.46 &34.91 &43.85 & 29.06 &24.23 &24.69 &19.73 & 5.85 \\

Gemini-3-flash
& 2 FPS
&\secondbest{}70.43 &36.79 &\thirdbest{}47.81 & 29.69 &30.84 &\secondbest{}45.31 &\secondbest{}31.10 & \secondbest{}16.72 \\

Gemini-3-flash text
& --
&51.68 &31.45 &38.44 &15.00 & \thirdbest{}32.60 &27.81 &19.73 & 0.25 \\

\specialrule{.2em}{.1em}{.1em}
\multicolumn{10}{c}{\textit{\bf Open-Source MLLM}} \\
\specialrule{.2em}{.1em}{.1em}

VideoGPT+ 4B
& 16
&56.01 &30.79 &\secondbest{}49.06 &18.79 &35.24 &17.87 &28.38 &4.72\\

Qwen3-VL 4B
& 32
& 58.99&\thirdbest{}40.88 &\secondbest{}49.06 &\secondbest{}32.19 &28.63 &25.62 &22.74 &9.39\\

Video-LLaVA 7B
& 8
&59.37 &35.78 &44.34 &14.43 &27.43 &22.78 &23.79 &3.48 \\

LLaVA-NeXT-Video-7B
& 32
&47.28 &34.08 &39.34 &22.50 &23.89 &24.44 &25.83 &0.34\\

InternVL3\_5-8B
& 8
&28.62 &31.03 &31.03 &31.13 &26.42 &\thirdbest{}28.62 &\thirdbest{}29.25 &3.27 \\

Qwen3-VL 8B
& 32
&62.36 &\secondbest{}41.82 &45.94 &\thirdbest{}31.87 &\secondbest{}37.89 &27.19 &23.08 & \thirdbest{}11.83 \\

\specialrule{.2em}{.1em}{.1em}
\multicolumn{10}{c}{\textit{\bf Motion-Specialized Model}} \\
\specialrule{.2em}{.1em}{.1em}

MotionLLM 7B
& 8
&46.97 &25.32 &39.62 &13.29 &19.38 &20.44 &18.64 &-6.47\\

\midrule
\midrule

{\bf Qwen3-VL 8B SFT (Ours)}
& 32
&\best{}87.26 &\best{}53.46 &\best{}68.44 &\best{}78.75 &\best{}38.77 &\best{}61.56 &\best{}52.17 &\best{}47.48\\

\specialrule{.2em}{.1em}{.1em}
\end{tabular}
    }
\end{table}
\begin{figure*}[t]

    \centering

    \includegraphics[width=\textwidth]{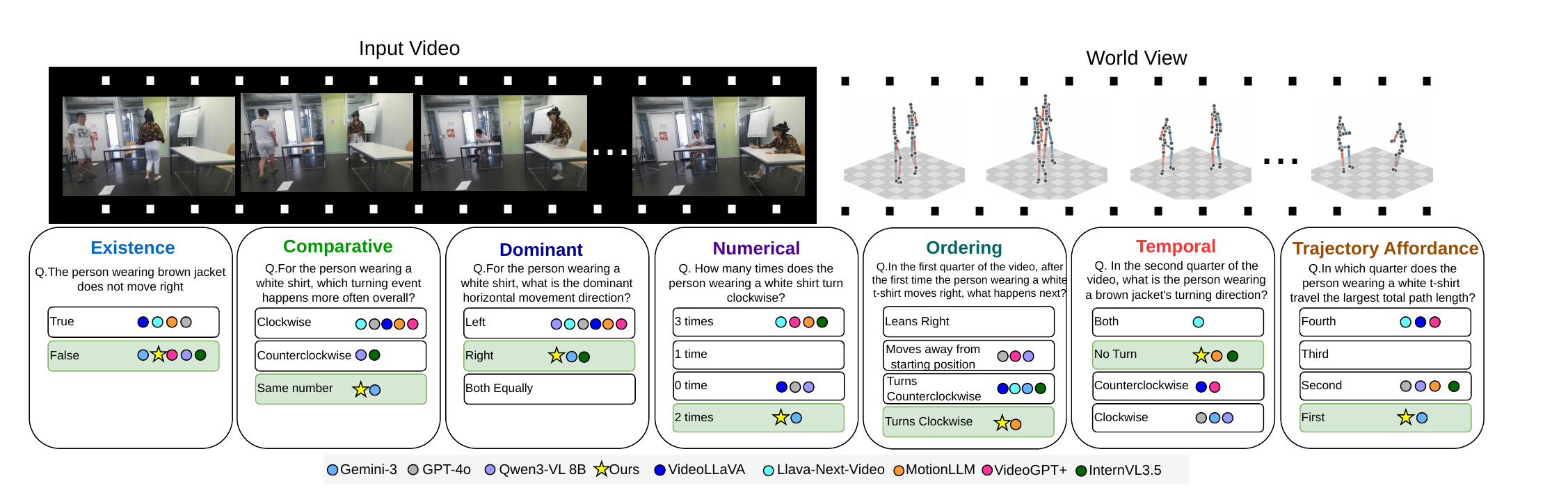}
    \includegraphics[width=\textwidth]{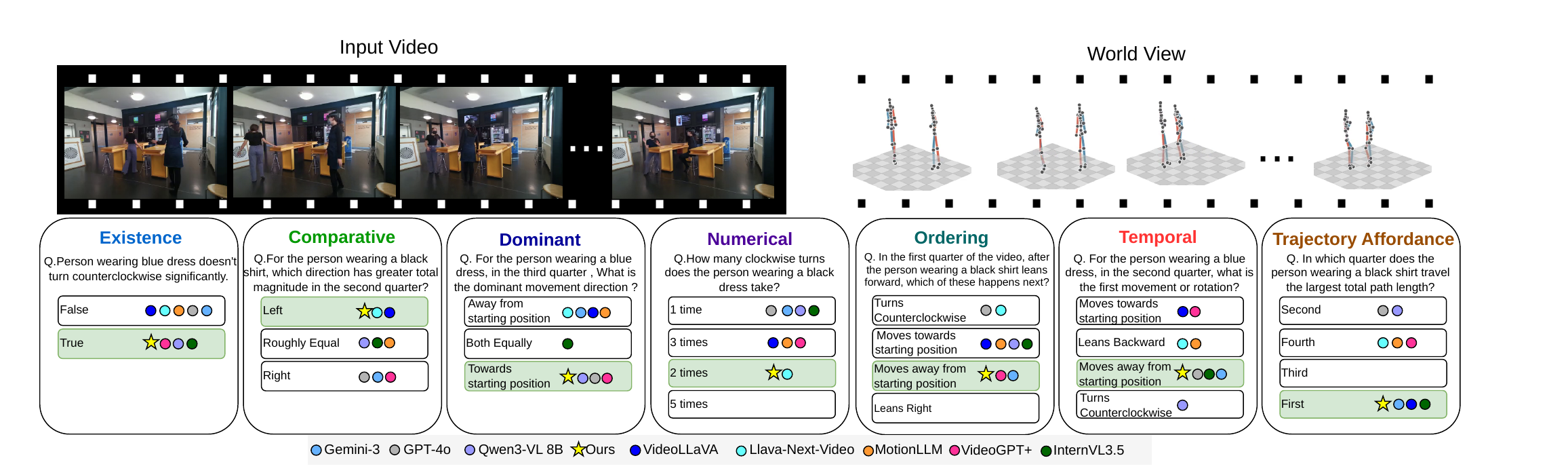}

    \caption{Qualitative results on the \textbf{EgoBody} dataset. The world-view visualization depicts extracted 3D SMPL-X poses and is shown for illustration only (not provided to the models). Green denotes the correct option. We compare predictions from multiple MLLMs across seven reasoning categories, where our model demonstrates more accurate and consistent responses.}
    
    \label{fig:egobody}


\end{figure*}

Among open-source models, VideoGPT+ performs well in Ordering ($35.24\%$), while Qwen3-VL 8B leads in Comparative and Dominant reasoning. Video-LLaVA,InternVL3.5 and LLaVA-Video-Next struggle, scoring near chance across most categories. MotionLLM exhibits a consistent failure in trajectory-based reasoning, with a Score of $-6.5$ and results significantly below chance in Numerical and Ordering tasks. Our Qwen3-VL 8B SFT model achieves the highest aggregate Score of $46.28$, with notable gains in Comparative (55.04), Dominant (67.19), Numerical (72.81) and Trajectory Affordance (52.17). Qualitative results are illustrated in Figure \ref{fig:rich}.

\noindent{\textbf{EgoBody --}} The EgoBody consists of static multi-view recordings of two interacting individuals in sequences ranging from 30-60 seconds. On this dataset, we only evaluate individual motion rather than social interactions to maintain consistency with our single-person trajectory focus; however, the presence of a second subject introduces significant complexity for localization and tracking. 
\begin{table}[h]
\centering
\caption{Evaluation of video MLLM models on the \textbf{EgoBody} split. Results are reported across seven reasoning categories and an overall normalized score. \best{Green}, \secondbest{orange}, and \thirdbest{blue} indicate the best, second-best, and third-best results per column, respectively.}
 \label{tab:egobody}
    \renewcommand{\arraystretch}{1.3}
    \setlength{\tabcolsep}{5pt}
    \resizebox{\textwidth}{!}{
    \begin{tabular}{l|c|c|c|c|c|c|c|c|c}
    \specialrule{.2em}{.1em}{.1em}
Model 
& Frames
& Existence 
& Comparative  
& Dominant 
& Numerical 
& Ordering 
& Temporal 
& Traj. Afford.
& Score \\

\specialrule{.2em}{.1em}{.1em}
Random Chance
& --
& 50.00
& 33.33
& 33.33
& 25.00
& 25.00
& 25.00
& 25.00 &
0.00\\
Human (subset)
& --
& 100.00
& 76.33
& 83.67
& 85.00
& 85.00
& 90.00
& 75.00 &
79.05\\
\specialrule{.2em}{.1em}{.1em}
\multicolumn{10}{c}{\textit{\bf Proprietary Models}} \\
\specialrule{.2em}{.1em}{.1em}

GPT-4o
& 32
&54.50 &30.39 &45.07 &26.79 &27.29 &29.34 &25.83 &4.93 \\

Gemini-3-flash
& 2 FPS
&\secondbest{}61.90 &32.38 &41.30 &\secondbest{}29.72 &27.13 &\thirdbest{}36.56 &\secondbest{}32.29 &\thirdbest{}9.80  \\

Gemini-3-flash text
& --
&55.95 &33.81 &42.26 &16.13 &\thirdbest{}33.13 &30.67 &24.27 & 4.52 \\

\specialrule{.2em}{.1em}{.1em}

\multicolumn{10}{c}{\textit{\bf Open-Source MLLM}} \\
\specialrule{.2em}{.1em}{.1em}

VideoGPT+ 4B
& 16
&48.51 &35.22 &49.66 &20.64 &23.21 &32.60 &\thirdbest{}27.54 &4.36 \\

Qwen3-VL 4B
& 32
&56.35 &36.88 &49.14 &\thirdbest{}27.55 &\secondbest{}33.99 &30.77 &25.00 & 9.31 \\

Video-LLaVA 7B
& 8
&58.24 &35.58 &\thirdbest{}51.00 &16.60 &24.89 &22.55 &22.91 &4.28\\

LLaVA-NeXT-Video-7B
& 32
&51.40 &\thirdbest{}39.01 &42.72 &24.26 &25.69 &24.71 &25.80 &3.84\\

InternVL3\_5-8B
& 8
&\thirdbest{}58.79 &36.98 &41.46 &25.57 &30.34 &\secondbest{}40.27 &25.21 &9.10 \\

Qwen3-VL 8B
& 32
&54.35 &\secondbest{}41.48 &\secondbest{}53.44 &27.24 &32.47 &32.00 &24.48 &\secondbest{}10.58\\

\specialrule{.2em}{.1em}{.1em}
\multicolumn{10}{c}{\textit{\bf Motion-Specialized Model}} \\
\specialrule{.2em}{.1em}{.1em}

MotionLLM 7B
& 8
&41.94 &24.78 &41.63 &9.84 &15.88 &20.10 &13.15&-10.03\\

\midrule
\midrule

{\bf QwenVL3-8B SFT (Ours)}
& 32
&\best{}79.32 &\best{}57.28 &\best{}65.30 &\best{}78.49 &\best{}31.25 &\best{}57.74 &\best{}52.50 &\best{}43.74\\
 
\specialrule{.2em}{.1em}{.1em}
\end{tabular}
    }
\end{table}
\renewcommand{\arraystretch}{1.15}
\setlength{\tabcolsep}{4pt}


As shown in Table \ref{tab:egobody}, Gemini-3-Flash drops to second place among zero-shot models, with its decreased performance in the Temporal category suggesting that multi-person environments complicate motion reasoning. GPT-4o performs poorly, while the text-only baseline achieves only half the accuracy of the video-text setting, confirming that the benchmark necessitates visual information.

\begin{figure*}[h]

    \centering

    \includegraphics[width=\textwidth]{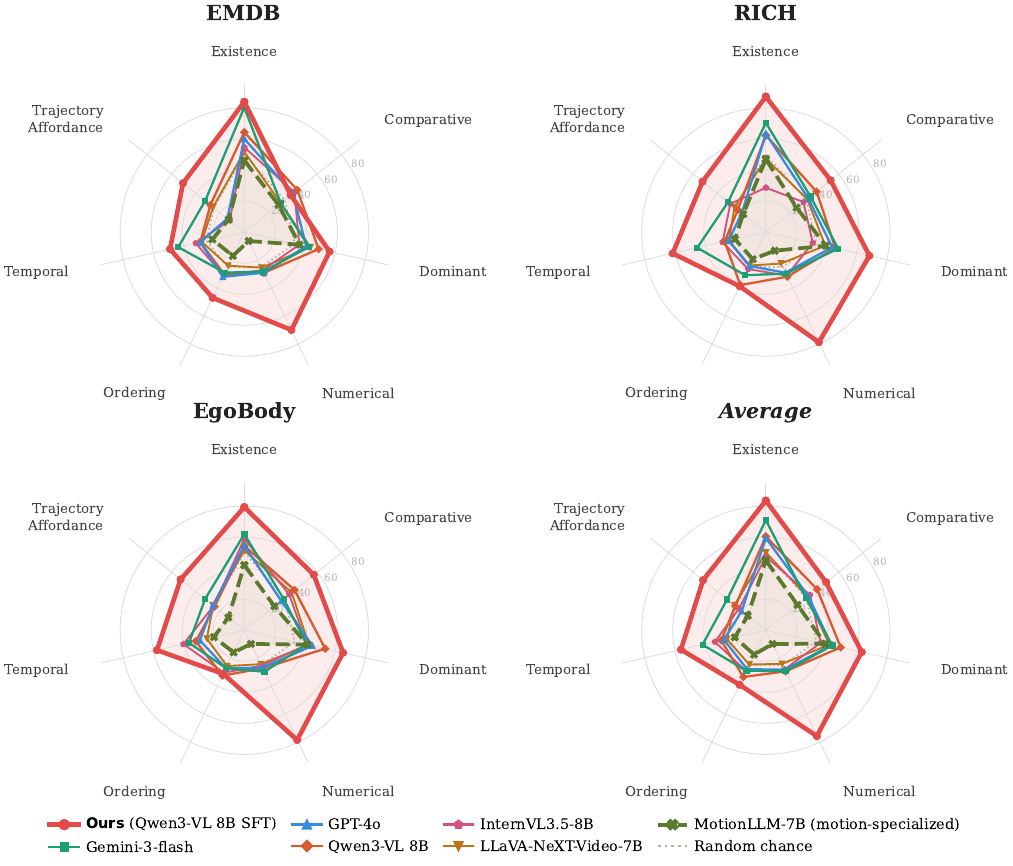}

    \caption{Performance comparison across reasoning categories on the \bench benchmark. Radar plots show model accuracy on seven categories for each dataset (EMDB, RICH, EgoBody) and their average. Our model (Qwen3-VL 8B SFT) consistently outperforms baselines across most categories, with particularly strong gains in numerical, temporal, and trajectory reasoning.}
    
    \label{fig:radar}


\end{figure*}

Open-source models generally struggle in this setting, though Qwen3-VL 8B remains the top baseline (Score: 10.58) with leading scores in Comparative and Dominant reasoning.InternVL3.5 performs well despite processing only 8 frames, achieving strong accuracy in Temporal (40.27) and Existence (58.79) categories.  VideoLLaVA shows localized strength in the Dominant category (51.00), likely reflecting dataset biases where participants frequently face one another. MotionLLM remains the worst performer (Score: $-10.03$), with Numerical and Ordering accuracies falling significantly below. Our Qwen3-VL 8B SFT model quadruples the base model's performance to an aggregate Score of 41.35, with substantial gains in Numerical and Trajectory Affordance tasks. However its low score on Ordering reflects the task being difficult in multi-person setting. Figure \ref{fig:egobody} demonstrates qualitative results. Figure \ref{fig:radar} visualizes the results across datasets. We provide qualitative video results across all three datasets in the supplementary material.

\subsection{Ablation}
\begin{table}[h]
\centering
\caption{Effect of Frame Count. Performance of Qwen3-VL 8B fine-tuned (SFT) with varying numbers of input video frames.}
 \label{tab:frame_count}
\vspace{2mm}

    \renewcommand{\arraystretch}{1.3} 
    \setlength{\tabcolsep}{5pt} 
    \resizebox{\textwidth}{!}{ 
    \begin{tabular}{llcccccccc}
\specialrule{.2em}{.1em}{.1em}
Dataset & Frames 
& Existence 
& Comparative
& Dominant
& Numerical 
& Ordering  
& Temporal
& Traj. Afford. 
& Score\\
\specialrule{.2em}{.1em}{.1em}

\multirow{3}{*}{EMDB}
& 16 
& 90.21 & 40.00 & 40.91 & 68.18 & 47.00 & 52.73 & 39.78 & 35.05 \\
& 32 
& 83.92 & 38.18 & 56.36 & 70.00 & 47.00 & 49.09 & 50.54 & 37.88 \\
& 64 
& 87.41 & 29.09 & 53.64 & 74.55 & 41.00 & 50.91 & 46.24 & 35.60 \\

\specialrule{.2em}{.1em}{.1em}

\multirow{3}{*}{RICH}
& 16 
& 86.06 & 53.77 & 64.38 & 76.56 & 40.53 & 58.44 & 51.51 & 45.53 \\
& 32 
& 87.26 & 53.46 & 68.44 & 78.75 & 38.77 & 61.56 & 52.17 & 47.48 \\
& 64 
& 87.26 & 55.03 & 67.19 & 72.81 & 39.65 & 60.00 & 52.17 & 46.29 \\

\specialrule{.2em}{.1em}{.1em}

\multirow{3}{*}{EgoBody}
& 16 
& 77.72 & 55.65 & 65.30 & 77.92 & 32.77 & 57.83 & 51.15 & 42.35 \\
& 32 
& 79.32 & 57.28 & 64.91 & 78.49 & 31.25 & 60.97 & 52.50 & 43.74 \\
& 64 
& 79.25 & 55.84 & 62.43 & 76.42 & 29.27 & 57.74 & 51.77 & 41.36 \\

\specialrule{.2em}{.1em}{.1em}
\end{tabular}
    }

\end{table}
\renewcommand{\arraystretch}{1.15}
\setlength{\tabcolsep}{4pt}
\begin{figure*}[t]

    \centering

    \includegraphics[width=\textwidth]{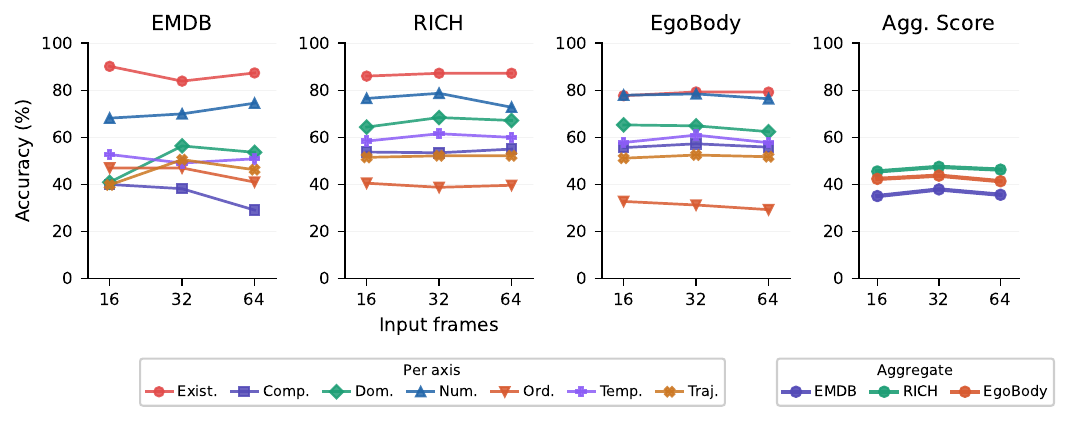}

    \vspace{-2mm}
    \caption{Effect of Frame Count. Qwen3-VL 8B is fine-tuned (SFT) with varying numbers of input video frames (16, 32, 64) and evaluated under matched settings. Results are reported on EMDB, RICH, and EgoBody across per-axis metrics and aggregate scores}
    
    \label{fig:framecount}


\end{figure*}
We investigate the following four questions here in the appendix:
\begin{enumerate}[topsep=0pt,partopsep=0pt,itemsep=0pt,parsep=0pt]
    \item How sensitive is the training to the number of frames?
    \item How sensitive is the training to the resolution of the videos?
    \item How does it perform if trained with a short reasoning trace?
    \item How does our model generalize to joint-level reasoning benchmarks like ActionArt?
\end{enumerate}
\noindent{\bf{Effect of Frame Count --}} We evaluated models trained with 16, 32, and 64 uniformly sampled frames to test the impact of temporal resolution. Results in Table \ref{tab:frame_count} indicate that performance is relatively stable, with 32 frames being optimal. On RICH and EgoBody, individual categories fluctuate by less than $3\pp$. On EMDB, however, Comparative performance declines while Numerical performance improves. Interestingly, Ordering shows a slight downward trend as frame counts increase, suggesting that excessive frames may introduce redundant information that hinders sequential reasoning. Figure \ref{fig:framecount} outlines these performances. 

\begin{table}[h]
\caption{Effect of input resolution on performance across datasets.}
\label{tab:resolution_ablation}
\centering

\renewcommand{\arraystretch}{1.2}
\setlength{\tabcolsep}{5pt}
\resizebox{\textwidth}{!}{
\begin{tabular}{llcccccccc}
\specialrule{.2em}{.1em}{.1em}
Dataset & Resolution
& Existence 
& Comparative  
& Dominant 
& Numerical 
& Ordering 
& Temporal 
& Traj. Afford.
& Score \\
\specialrule{.2em}{.1em}{.1em}

\multirow{3}{*}{EMDB}
& 128$\times$128 
&80.42 &38.18 &51.82 &69.09 &36.00 &50.00 &33.33 &30.53 \\
& 256$\times$256 
&83.92 &38.18 &56.36 &70.00 &47.00 &49.09 &50.54 &37.88 \\
& 384$\times$384 
&82.52 &31.82 &54.55 &62.73 &39.00 &51.82 &43.01 &31.91 \\
\specialrule{.2em}{.1em}{.1em}

\multirow{3}{*}{RICH}
& 128$\times$128 
&77.88 &50.00 &48.44 &77.19 &34.80 &49.69 &43.48 &34.81 \\
& 256$\times$256 
&87.26 &53.46 &68.44 &78.75 &38.77 &61.56 &52.17 &47.48 \\
& 384$\times$384 
&85.49 &52.28 &58.68 &75.94 &29.52 &54.69 &46.47 &40.94 \\

\specialrule{.2em}{.1em}{.1em}

\multirow{3}{*}{EgoBody}
& 128$\times$128 
&73.88 &52.49 &57.74 &74.34 &32.77 &53.94 &48.96 &37.11 \\
& 256$\times$256 
&79.32 &57.28 &64.91 &78.49 &31.25 &60.97 &52.50 &43.74 \\
& 384$\times$384 
&76.11 &54.67 &55.45 &74.83 &30.12 &52.61 &48.02 &36.88 \\

\specialrule{.2em}{.1em}{.1em}
\end{tabular}
}
\end{table}
\begin{figure*}[t]

    \centering

    \includegraphics[width=\textwidth]{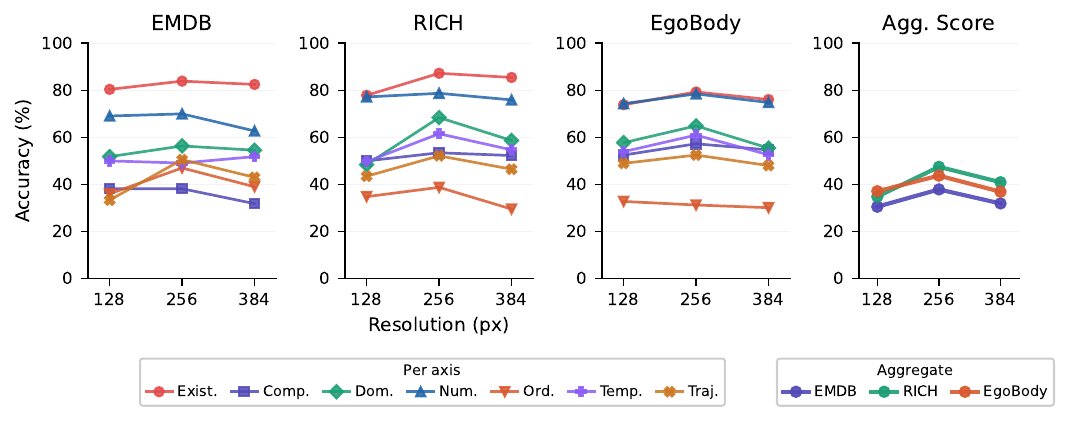}

    \vspace{-2mm}
    \caption{Effect of input resolution. Qwen3-VL 8B is fine-tuned (SFT) at different input resolutions and evaluated at matched resolutions. Results are reported on EMDB, RICH, and EgoBody across per-axis metrics and overall aggregate scores.}
    
    \label{fig:resolution}


\end{figure*}
\noindent{\bf{Effect of Resolution --}} We ablated input resolutions ($128 \times 128$, $256 \times 256$, $384\times384$) at 32 frames.  Table ~\ref{tab:resolution_ablation}
shows per-axis accuracy and aggregate
Score across all three datasets. 

All datasets exhibit an "inverted V" pattern, with $256 \times 256$ being consistently optimal. Both lower and higher resolutions degrade performance, though the drop from $256 \times 256$ to $384\times 384$ is more pronounced than from $128\times 128$ to $256\times256$. Higher resolution likely dilutes the motion signal by amplifying background tokens, an effect most pronounced in RICH where subjects are distant from the camera.

While Existence, Numerical, and Comparative tasks remain stable due to their reliance on detection, Trajectory Affordance suffers at low resolutions ($128\times128$) which requires finer spatial resolution to track positional changes. Figure \ref{fig:resolution}, indicate $256 \times 256$ is the optimal resolution, balancing motion signal preservation with the prevention of background token dilution.

\noindent{\bf{Reasoning Supervision --}} We tested whether adding a short reasoning trace during SFT improves performance. To do this, the SFT Reason is trained with the same training samples and LoRA configuration. However we add a short reasoning trace to the training samples using the same \textit{SpatialCodes} that we have. It follows the template of <reasoning></reasoning><answer><answer> and the reasoning trace has structured keys describing the relevant motion before coming to an answer.
\begin{table}[h]
\centering
\caption{Reasoning supervision. Results comparing SFT fine-tuning with answer options only and with short reasoning traces.}
\label{tab:reasoning}
\vspace{2mm}
\renewcommand{\arraystretch}{1.2}
\setlength{\tabcolsep}{4pt}
\resizebox{\textwidth}{!}{
\begin{tabular}{llcccccccc}
\specialrule{.2em}{.1em}{.1em}
Dataset & Method 
& Existence & Comparative & Dominant & Numerical & Ordering & Temporal & Traj. Afford. & Score \\
\specialrule{.2em}{.1em}{.1em}

\multirow{2}{*}{EMDB}
& SFT 
& 83.92 & 38.18 & 56.36 & 70.00 & 47.00 & 49.09 & 50.54 & 37.88 \\

& SFT Reason
& 80.15 & 37.86 & 55.14 & 75.70 & 36.86 & 54.21 & 33.33 & 32.58 \\

\specialrule{.2em}{.1em}{.1em}

\multirow{2}{*}{RICH}
& SFT
& 87.26 & 55.03 & 67.19 & 72.81 & 39.65 & 60.00 & 52.17 & 46.28 \\

& SFT Reason
& 81.15 & 54.84 & 55.93 & 70.88 & 29.20 & 57.00 & 46.69 & 38.11 \\

\specialrule{.2em}{.1em}{.1em}

\multirow{2}{*}{EgoBody}
& SFT 
& 79.25 & 55.84 & 62.43 & 76.42 & 29.27 & 57.74 & 51.77 & 41.35 \\

& SFT Reason
& 77.69 & 51.20 & 55.24 & 73.70 & 27.72 & 49.28 & 45.34 & 34.60 \\

\specialrule{.2em}{.1em}{.1em}
\end{tabular}
}
\end{table}
Table \ref{tab:reasoning} shows that this "SFT Reason" model underperforms the option-only SFT across all categories. The drop is most severe in Ordering ($-10.5\pp$ on RICH), Trajectory Affordance ( $-17.2\pp$ on EMDB), and Dominant ($-11.3\pp$ on RICH) as the added complexity of the reasoning trace leads to compounding errors in multi-step inference.
Figure \ref{tab:reasoning} shows examples of output reasoning traces. The model was trained with similar reasoning traces.

\noindent{\bf{Out-of-Domain Evaluation --}} We evaluated our Qwen3-VL 8B SFT model (trained on \bench) on the ActionArt~\cite{ActionArt} benchmark, which is the closest existing benchmark with similar categories such as Action Direction (AD), Action Count (AC), Action Sequence (AS), Global Spatial (GS), Local Spatial (LS), Temporal Localization (TL) among others but focuses on joint-level pose reasoning. The results are shown in Table \ref{actionart}

\begin{table}[h]

    \centering
    \caption{Evaluation on the ActionArt benchmark. The ActionArt score is taken from the original paper, while all other results are computed in our evaluation. Column names follow the ActionArt convention.}
    \label{actionart}
    \renewcommand{\arraystretch}{1.3}
    \setlength{\tabcolsep}{5pt}
    \resizebox{\textwidth}{!}{
    \begin{tabular}{lccccccccc}
\specialrule{.2em}{.1em}{.1em}
Model 
& AC 
& MV 
& AS 
& HO 
& AR
& GS 
& LS 
& TL 
& Avg \\
\midrule
VideoLLaVA
& 33.33  
& 37.33  
& 37.93  
& 55.68  
& 38.67  
& 62.18  
& 34.22  
& 36.28  
& 41.95  
\\
MotionLLM
& 24.60  
& 32.96  
& 27.27  
& 28.09  
& 23.47  
& 52.30  
& 21.37  
& 21.58  
& 28.96  
\\
VideoGPT+
& 37.30  
& 44.44  
& 44.03  
& 59.55  
& 56.59  
& 61.51  
& 52.73  
& 31.71  
& 48.48  

\\
ActionArt
& 45.20  
& 59.60  
& 61.80  
& 83.10  
& 82.60  
& 85.50  
& 75.70  
& 42.50  
& 69.40  

\\
Qwen3VL-8B
& 52.80 
& 48.74 
& 53.35 
& 72.29 
& 71.57 
& 83.40 
& 66.10 
& 39.08 
& 60.92 \\
Qwen3VL-8B SFT
& 40.00  
& 39.50  
& 44.41  
& 46.99  
& 43.46  
& 77.55  
& 43.44  
& 37.54  
& 46.61  
\\
\specialrule{.2em}{.1em}{.1em}
\end{tabular}
    }

\end{table}

\renewcommand{\arraystretch}{1.15}
\setlength{\tabcolsep}{4pt}
Our model scored 46.61, lower than the zero-shot Qwen3-VL 8B baseline. This suggests that trajectory-level supervision does not substitute for fine-grained joint-pose information. However, our model significantly outperformed MotionLLM (28.96). 
We conclude that trajectory-level and joint-level understanding are complementary skills requiring distinct supervision signals. 
Our SFT teaches the model to reason about whole-body displacement, orientation, and temporal sequencing, but this does not substitute for the fine-grained joint-pose supervision that ActionArt demands.

\subsection{Broader Impact}
This work introduces a benchmark for evaluating human trajectory and orientation reasoning with Video MLLMs. The dataset and benchmark can support the development of downstream models for applications such as sports analytics and industrial safety. However, models evaluated on this dataset could potentially be misused for human surveillance or behavior analysis in sensitive contexts. To mitigate this, the dataset only provides annotations and references to source data, requiring users to obtain the original datasets and comply with their respective licenses.

\begin{table}[h]
\centering
\caption{Per-category vs.\ joint training. Performance of Qwen3-VL 8B fine-tuned (SFT) on individual categories and evaluated across datasets. Green, orange, and blue denote the best, second-best, and third-best results per dataset, respectively. Zero-shot indicates no task-specific training, while All denotes joint training on all categories.}
\label{tab:axis_ablation_all_datasets}
\renewcommand{\arraystretch}{1.2}
\setlength{\tabcolsep}{4pt}
\resizebox{\textwidth}{!}{
\begin{tabular}{llcccccccc}
\specialrule{.2em}{.1em}{.1em}
Dataset & Setting
& Existence 
& Comparative  
& Dominant 
& Numerical 
& Ordering 
& Temporal 
& Traj. Afford.
& Score \\
\specialrule{.2em}{.1em}{.1em}

\multirow{7}{*}{EMDB}
& Zero-shot
&64.34 &\best43.64 &\secondbest49.09 &29.09 &30.00 &29.09 &27.96 &12.83 \\

& Ordering
&72.73 &40.00 &43.64 &33.64 &\secondbest40.00 &\thirdbest30.91 &\thirdbest29.03 &\thirdbest16.52 \\

& Numerical
&70.63 &\secondbest40.91 &\thirdbest47.27 &\best70.00 &\thirdbest36.00 &26.36 &17.20 &\secondbest19.94 \\

& Comp.+Dom.
&74.83 &38.18 &\best56.36 &\secondbest37.27 &26.00 &26.36 &24.73 &15.80 \\

& Temporal
&\secondbest79.72 &30.91 &39.09 &35.45 &28.00 &\best50.00 &20.43 &15.66 \\

& Traj. Afford.
&\thirdbest75.00 &\thirdbest39.09 &40.00 &\thirdbest36.36 &28.00 &21.82 &\secondbest43.01 &15.60 \\

& All
&\best83.92 &38.18 &\best56.36 &\best70.00 &\best47.00 &\secondbest49.09 &\best50.54 &\best37.88 \\

\specialrule{.2em}{.1em}{.1em}

\multirow{7}{*}{RICH}
& Zero-shot
&58.99 &40.88 &49.06 &32.19 &28.63 &25.62 &22.74 &9.39 \\

& Ordering
&56.97 &35.85 &47.81 &29.38 &\secondbest35.24 &25.56 &28.76 &9.43 \\

& Numerical
&67.79 &\thirdbest41.96 &50.00 &\secondbest76.56 &\thirdbest30.84 &\thirdbest33.44 &21.74 &\secondbest22.26 \\

& Comp.+Dom.
&68.75 &\secondbest51.57 &\secondbest66.88 &27.50 &25.99 &31.25 &\thirdbest30.77 &\thirdbest19.40 \\

& Temporal
&\secondbest75.00 &37.42 &43.75 &\thirdbest33.75 &27.75 &\secondbest56.56 &29.77 &19.36 \\

& Traj. Afford.
&\thirdbest72.60 &38.99 &\thirdbest56.25 &25.62 &20.26 &26.25 &\secondbest47.16 &16.25 \\

& All
&\best87.26 &\best53.46 &\best68.44 &\best78.75 &\best38.77 &\best61.56 &\best52.17 &\best47.48 \\

\specialrule{.2em}{.1em}{.1em}

\multirow{7}{*}{EgoBody}
& Zero-shot
&54.35 &41.48 &53.44 &27.24 &\secondbest32.47 &32.00 &24.48 &10.58 \\

& Ordering
&59.80 &42.15 &50.67 &26.70 &\thirdbest31.86 &31.43 &27.50 &11.73 \\

& Numerical
&\thirdbest66.84 &45.79 &55.45 &\secondbest77.83 &28.96 &\thirdbest35.33 &28.44 &\secondbest25.61 \\

& Comp.+Dom.
&62.99 &\secondbest55.94 &\secondbest58.99 &28.49 &25.76 &33.14 &\thirdbest32.40 &17.81 \\

& Temporal
&\secondbest71.99 &41.48 &45.79 &\thirdbest35.47 &27.44 &\secondbest54.61 &28.65 &\thirdbest19.49 \\

& Traj. Afford.
&65.09 &\thirdbest45.98 &\thirdbest55.64 &25.09 &24.24 &30.77 &\secondbest47.81 &17.11 \\

& All
&\best79.32 &\best57.28 &\best64.91 &\best78.49 &\thirdbest31.25 &\best60.97 &\best52.50 &\best43.74 \\

\specialrule{.2em}{.1em}{.1em}
\end{tabular}
}

\end{table}
\begin{longtable}{p{2.2cm} p{4.6cm} p{7.2cm}}
\caption{Overview of the seven categories, highlighting the capability tested in each, along with a representative example question.}
\label{tab:spatial_axes}\\
\toprule
\textbf{Category} & \textbf{Sub-axis} & \textbf{Example question} \\
\midrule
\endfirsthead

\toprule
\textbf{Category} & \textbf{Sub-axis} & \textbf{Example question} \\
\midrule
\endhead

\bottomrule
\endfoot

\multirow{3}{*}{existence}
& Semantic existence
& Is there walking in the video? \\

& Directional existence
& Does the person move to the left at any point? \\

& Negation (logical complement)
& The person did not move to the left at any point. True or false? \\
\midrule
\multirow{3}{*}{comparative}
& Count-based comparison
& Which event happens more often overall: moving left or moving right? \\

& Magnitude-based comparison
& Which direction has greater total magnitude overall: left or right? \\

& Speed-based comparison
& Is the person's first move left faster or slower than their first move right? \\
\midrule

\multirow{2}{*}{dominant}
& Direction dominance (global)
& What is the dominant horizontal movement direction, left or right? \\

& Direction dominance (quarter-local)
& In the second quarter of the video, what is the primary turning direction, clockwise or counterclockwise? \\
\midrule

\multirow{3}{*}{numerical}
& Event count (direction-specific)
& How many times does the person move to the left? \\

& Total magnitude (direction-specific)
& How far does the person move to the left in total? \\

& Strength-restricted count (moderate/significant)
& How many times does the person turn clockwise significantly? \\
\midrule

ordering
& Next event after anchor within quarter
& In the first quarter of the video, after the person moves left, which of these happens next? \\
\midrule

\multirow{3}{*}{temporal}
& Quarter-local motion
& In the first quarter of the video, what horizontal movement does the person make? \\

& First event in quarter
& In the second quarter of the video, what is the first movement or rotation that happens? \\

& Event speed classification
& During the person's first move to the left, how fast is the movement? \\
\midrule

\multirow{5}{*}{\shortstack{trajectory\\affordance}}
& Largest net displacement (quarter)
& In which quarter does the person move most to the right (largest net rightward displacement)? \\

& Smallest net displacement (quarter)
& In which quarter does the person move least to the right (smallest non-zero net rightward displacement)? \\

& Largest path length (quarter)
& In which quarter does the person travel the largest total path length? \\

& Overall endpoint position
& Where is the person horizontally relative to where they started at the end of the video? \\

& Trajectory shape
& How does the person's overall position evolve relative to their starting position? \\

\end{longtable}

\begin{figure}[t]
\centering
\begin{tcolorbox}[colback=gray!5, colframe=black, width=0.92\linewidth, boxrule=0.5pt]
\textbf{System Prompt}

\small
You are answering multiple-choice questions about a person's motion in a video.

\medskip
\textbf{Coordinate System}
\begin{itemize}
    \item Motion is measured relative to the first frame of the video.
    \item The vertical axis (\textbf{Y}) points upward from the ground.
    \item The horizontal axis (\textbf{X}) points left--right relative to the person in the first frame.
    \item The depth axis (\textbf{Z}) points forward--backward relative to the person in the first frame.
    \item Turning left or right refers to rotation of the person's body, not camera motion.
\end{itemize}

\medskip
\textbf{Temporal Segmentation}
When a question refers to quarters of the video, these correspond to four equal-length segments:
first quarter (0--25\%), second (25--50\%), third (50--75\%), and fourth (75--100\%).

\medskip
\textbf{Answer Format}
Select the single best option from the provided choices. Respond with the option letter and its corresponding text.
\end{tcolorbox}
\caption{System prompt used for evaluation of models. We use the same system prompt for training SFT.}
\label{fig:system_prompt}
\end{figure}
\begin{figure}[t]
\centering
\fbox{
\begin{minipage}{0.95\linewidth}
\small
\textbf{Clothing Captioning Prompt}

\medskip
Describe only the person's clothing and visible accessories. Mention garments, colors, patterns, materials, shoes, hats, glasses, bags, or jewelry if visible. Do not mention the person, pose, action, body position, camera view, background, skateboard, or scene. Return one short clothing-only phrase, not a sentence.
\end{minipage}
}
\caption{\normalsize Prompt used for clothing caption generation.}
\label{fig:cloth_prompt}
\end{figure}
\begin{figure}[t]
\centering
\begin{tcolorbox}[
    colback=gray!5,
    colframe=black,
    width=0.96\linewidth,
    boxrule=0.5pt,
    title=\textbf{Reasoning traces by category}
]
\small

\textbf{Existence}

\textit{Question:} Does the person move to the left at any point?  
\textit{Options:} A. Yes \quad B. No

\texttt{<reasoning>target\_event\_count=7; choice=Yes; option=A</reasoning>}\\
\texttt{<answer>A</answer>}

\textit{Ground truth:} A

\medskip
\hrule
\medskip

\textbf{Comparative}

\textit{Question:} Which event happens more often in the second quarter of the video: moving right or moving left?  
\textit{Options:} A. same number of events \quad B. right \quad C. left

\texttt{<reasoning>left\_count=2; right\_count=1; choice=left; option=C</reasoning>}\\
\texttt{<answer>C</answer>}

\textit{Ground truth:} C

\medskip
\hrule
\medskip

\textbf{Dominant}

\textit{Question:} What is the dominant depth movement direction, towards or away from the starting position?  
\textit{Options:} A. both towards and away equally \quad B. towards starting position \quad C. away from starting position

\texttt{<reasoning>away=42.0; towards=10.0; diff=32.0; tie\_threshold=8.4; choice=away from starting position; option=C</reasoning>}\\
\texttt{<answer>C</answer>}

\textit{Ground truth:} C

\medskip
\hrule
\medskip

\textbf{Numerical}

\textit{Question:} How many times does the person lean to the left moderately?  
\textit{Options:} A. 0 \quad B. 3 \quad C. 2 \quad D. 1

\texttt{<reasoning>count=0; choice=0; option=A</reasoning>}\\
\texttt{<answer>A</answer>}

\textit{Ground truth:} A

\medskip
\hrule
\medskip

\textbf{Ordering}

\textit{Question:} In the second quarter of the video, after the first time the person leans to the right, which of these happens next?  
\textit{Options:} A. moves away from starting position \quad B. turns clockwise \quad C. moves towards starting position \quad D. leans to the left

{\ttfamily\small{<reasoning>anchor=rotation\_roll(-4.0,0.15,significant\_leaning\_right,moderate); next\_event=rotation\_roll(4.0,0.29,significant\_leaning\_left,moderate); choice=leans to the left; option=D</reasoning>}\\}
\texttt{<answer>D</answer>}

\textit{Ground truth:} D

\medskip
\hrule
\medskip

\textbf{Temporal}

\textit{Question:} In the fourth quarter of the video, what is the first movement or rotation that happens?  
\textit{Options:} A. moves away from starting position \quad B. leans backward \quad C. leans to the right \quad D. moves right

{\ttfamily\small

\textless reasoning\textgreater presence=first;

anchor=rotation\_pitch(-8.0,0.14,significant\_leaning\_backward,moderate); choice=leans backward; option=B\textless/reasoning\textgreater

}
\texttt{<answer>B</answer>}

\textit{Ground truth:} B

\medskip
\hrule
\medskip

\textbf{Trajectory Affordance}

\textit{Question:} How does the person's overall position evolve relative to their starting position?  
\textit{Options:} A. returns close to starting position by the end \quad B. oscillates but ends far from start \quad C. remains near starting position throughout \quad D. moves away and stays away

\texttt{<reasoning>direction\_changes=18; final\_dist\_units=40.86; max\_dist\_units=45.48; return\_threshold\_units=4.55; shape=returns\_close\_to\_starting\_position\_by\_end; choice=returns close to starting position by the end; option=A</reasoning>}\\
\texttt{<answer>A</answer>}

\textit{Ground truth:} A

\end{tcolorbox}
\caption{Examples of category-wise reasoning traces for a sample EMDB video. For each category, we show the question, answer choices, model reasoning trace, prediction, and ground-truth answer.}
\label{fig:reasoning_traces}
\end{figure}
\begin{longtable}{p{4cm} p{6cm} p{4cm}}
\caption{ Spatial Codes and corresponding discretization into categorical bins. Translation is measured in units ($1$ unit = $10$\,cm), and rotation is measured in degrees.}
\label{tab:spatial_codes}\\

\toprule
\textbf{Axis} & \textbf{Category} & \textbf{Range} \\
\midrule
\endfirsthead

\toprule
\textbf{Axis} & \textbf{Category} & \textbf{Range} \\
\midrule
\endhead

\bottomrule
\endfoot

\multicolumn{3}{l}{\textbf{Translation (Displacement)}} \\
\midrule

\multirow{11}{*}{X-axis}
& very\_long\_left & $v < -10$ \\
& long\_left & $-10 \leq v < -8$ \\
& moderate\_left & $-8 \leq v < -5$ \\
& short\_left & $-5 \leq v < -3$ \\
& very\_short\_left & $-3 \leq v < -1$ \\
& no\_action & $-1 \leq v \leq 1$ \\
& very\_short\_right & $1 < v \leq 3$ \\
& short\_right & $3 < v \leq 5$ \\
& moderate\_right & $5 < v \leq 8$ \\
& long\_right & $8 < v \leq 10$ \\
& very\_long\_right & $v > 10$ \\

\midrule

\multirow{11}{*}{Y-axis}
& very\_long\_down & $v < -10$ \\
& long\_down & $-10 \leq v < -8$ \\
& moderate\_down & $-8 \leq v < -5$ \\
& short\_down & $-5 \leq v < -3$ \\
& very\_short\_down & $-3 \leq v < -1$ \\
& no\_action & $-1 \leq v \leq 1$ \\
& very\_short\_up & $1 < v \leq 3$ \\
& short\_up & $3 < v \leq 5$ \\
& moderate\_up & $5 < v \leq 8$ \\
& long\_up & $8 < v \leq 10$ \\
& very\_long\_up & $v > 10$ \\

\midrule

\multirow{11}{*}{Z-axis}
& very\_long\_backward & $v < -10$ \\
& long\_backward & $-10 \leq v < -8$ \\
& moderate\_backward & $-8 \leq v < -5$ \\
& short\_backward & $-5 \leq v < -3$ \\
& very\_short\_backward & $-3 \leq v < -1$ \\
& no\_action & $-1 \leq v \leq 1$ \\
& very\_short\_forward & $1 < v \leq 3$ \\
& short\_forward & $3 < v \leq 5$ \\
& moderate\_forward & $5 < v \leq 8$ \\
& long\_forward & $8 < v \leq 10$ \\
& very\_long\_forward & $v > 10$ \\

\midrule
\multicolumn{3}{l}{\textbf{Rotation (Orientation)}} \\
\midrule

\multirow{7}{*}{Pitch }
& significant\_leaning\_backward & $v < -4$ \\
& moderate\_leaning\_backward & $-4 \leq v < -3$ \\
& slight\_leaning\_backward & $-3 \leq v < -2$ \\
& no\_action & $-2 \leq v \leq 0$ \\
& slight\_leaning\_forward & $0 < v \leq 2$ \\
& moderate\_leaning\_forward & $2 < v \leq 3$ \\
& significant\_leaning\_forward & $v > 3$ \\

\midrule

\multirow{7}{*}{Roll}
& significant\_leaning\_right & $v < -4$ \\
& moderate\_leaning\_right & $-4 \leq v < -3$ \\
& slight\_leaning\_right & $-3 \leq v < -2$ \\
& no\_action & $-2 \leq v \leq 0$ \\
& slight\_leaning\_left & $0 < v \leq 2$ \\
& moderate\_leaning\_left & $2 < v \leq 3$ \\
& significant\_leaning\_left & $v > 3$ \\

\midrule

\multirow{7}{*}{Yaw }
& significant\_turn\_clockwise & $v < -4$ \\
& moderate\_turn\_clockwise & $-4 \leq v < -3$ \\
& slight\_turn\_clockwise & $-3 \leq v < -2$ \\
& no\_action & $-2 \leq v \leq 0$ \\
& slight\_turn\_counterclockwise & $0 < v \leq 2$ \\
& moderate\_turn\_counterclockwise & $2 < v \leq 3$ \\
& significant\_turn\_counterclockwise & $v > 3$ \\

\end{longtable}
\begin{table}[t]
\centering
\caption{Spatial codes discretization for temporal categories based on velocity magnitude. Velocity is measured in units/frame; at 30 FPS, multiply by 30 to obtain units/second.}
\label{tab:temporal_codes}

\small
\setlength{\tabcolsep}{8pt}
\renewcommand{\arraystretch}{1.15}
\begin{tabular}{p{5cm} p{5cm}}
\toprule
\textbf{Temporal category} & \textbf{Velocity range} \\
\midrule
very\_slow & $|v| \leq 0.05$ \\
slow & $0.05 < |v| \leq 0.1$ \\
moderate & $0.1 < |v| \leq 0.5$ \\
fast & $0.5 < |v| \leq 0.8$ \\
very\_fast & $|v| > 0.8$ \\
\bottomrule
\end{tabular}

\end{table}
\clearpage
\end{document}